\documentclass[opre,nonblindrev]{informs3}

\OneAndAHalfSpacedXI 

\usepackage{natbib}
 \bibpunct[, ]{(}{)}{,}{a}{}{,}%
 \def\bibfont{\small}%

\usepackage{algorithm}
\usepackage{algorithmic}
\usepackage{color}
\usepackage{hyperref}
\usepackage{times,amsmath,amsfonts,amssymb}
\usepackage{amsfonts}
\usepackage{comment}
\usepackage{booktabs}
\usepackage{threeparttable}
\usepackage{subcaption}
\usepackage{graphicx}

\newcommand\bgnd{\textit{b}GND}
\newcommand\bgnds{\textit{b}GND }
\newcommand\E{\mathbb{E}}
\newcommand\Prob{P}
\newcommand\minprob{\rho_X}
\newcommand\Real{\mathbb{R}}
\newcommand\X{\mathcal{X}}
\newcommand\samp{\mathcal{S}}

\newcommand\e{\epsilon}
\newcommand\Fspace{\mathcal{F}}
\newcommand\muhat{\hat\mu}
\newcommand\mustar{\mu^\ast}
\newcommand\mulo{\underline{\mu}}
\newcommand\muhi{\overline{\mu}}
\newcommand\bhat{\hat b}
\newcommand\betahat{\hat\beta}
\newcommand\betastar{\beta^\ast_{\hat\mu}}
\newcommand\bstar{b^\ast_{\hat\mu}}
\newcommand\blo{\underline{b}}
\newcommand\bhi{\overline{b}}
\newcommand\mhat{\hat{m}}

\TheoremsNumberedThrough     
\ECRepeatTheorems

\EquationsNumberedThrough    

\MANUSCRIPTNO{}

\begin{document}


\RUNAUTHOR{G\"{u}rlek, de V\'ericourt, and Lee}

\RUNTITLE{Boosted Generalized Normal Distributions}

\TITLE{Boosted Generalized Normal Distributions: Integrating Machine Learning with Operations Knowledge}

\ARTICLEAUTHORS{%
\AUTHOR{Rag{\i}p G\"{u}rlek}
\AFF{Goizueta Business School, Emory University}

\AUTHOR{Francis de V\'ericourt}
\AFF{European School of Management and Technology Berlin}

\AUTHOR{Donald K.K. Lee\thanks{Corresponding author. Supported by National Institutes of Health grant R01-HL164405.}}
\AFF{Goizueta Business School and Department of Biostatistics \& Bioinformatics, Emory University}
} 

\ABSTRACT{
Applications of machine learning (ML) techniques to operational settings often face two challenges: i) ML methods mostly provide point predictions whereas many operational problems require distributional information; and ii) They typically do not incorporate the extensive body of knowledge in the operations literature, particularly the theoretical and empirical findings that characterize specific distributions.
We introduce a novel and rigorous methodology, the Boosted Generalized Normal Distribution (\bgnd), to address these challenges. The Generalized Normal Distribution (GND) encompasses a wide range of parametric distributions commonly encountered in operations, and \bgnds leverages gradient boosting with tree learners to flexibly estimate the parameters of the GND as functions of covariates.
We establish \bgnd's statistical consistency, thereby extending this key property to special cases studied in the ML literature that lacked such guarantees.
Using data from a large academic emergency department in the United States, we show that the distributional forecasting of patient wait and service times can be meaningfully improved by leveraging findings from the healthcare operations literature.
Specifically, \bgnds performs 6\% and 9\% better than the distribution-agnostic ML benchmark used to forecast wait and service times respectively. Further analysis suggests that these improvements translate into a 9\% increase in patient satisfaction and a 4\% reduction in mortality for myocardial infarction patients. Our work underscores the importance of integrating ML with operations knowledge to enhance distributional forecasts.
}%


\KEYWORDS{Distributional Machine Learning, Gradient Boosting, Wait Times, Service Times, Emergency Departments, Healthcare Operations} \HISTORY{1 August 2024}

\maketitle
%

\section{Introduction}
Machine learning (ML) is increasingly being implemented across various areas of operations,
including forecasting patient wait times in emergency departments \citep{arora2023probabilistic}, developing chemotherapy treatments \citep{bertsimasAnalytics2016}, manufacturing process improvements \citep{senonerUsing2022}, to optimizing last-mile delivery assignments \citep{liuOnTime2021}. The growing application of these methodologies can be attributed to their ability to learn complex relationships, their remarkable versatility, and the increased availability of operations-related data.

However, there are two challenges to applying ML techniques to operational settings. First, the majority of ML methods provide point predictions
rather than distributional forecasts. Yet, many fundamental operations problems are framed around probability distributions: The analysis of service systems, for instance, focus on the distributions of wait and service times, while the newsvendor problem requires a specific quantile of the demand distribution. Consider also staffing \citep{he2012timing, banBig2019} and elective surgery scheduling \citep{rathIntegrated2017}, which rely on distributional estimates of hospital workload and surgery duration.


Second, the distributional ML algorithms do not account for the extensive body of knowledge in operations \citep{arora2023probabilistic, bertsimas2022predicting}. This is because nonparametric ML algorithms are distribution-agnostic by design, precluding them from taking advantage of the specific distributional knowledge identified in the theoretical and empirical operations literature. For instance, \cite{kingman1962} demonstrated that wait times in queuing systems under heavy traffic can be approximated by an exponential distribution, and empirical evidence suggests this holds even in less heavily loaded systems \citep{brown2005statistical}. The empirical queuing literature  also established that service duration in various  operations settings can be  approximated by a log-normal distribution \citep{brown2005statistical,he2012timing,armony2015patient,ding2024nurse}. The distributional ML algorithms currently used in operational contexts overlook this knowledge, which can potentially be exploited for performance gains.

In this paper, we introduce a novel distributional ML approach called the \emph{boosted Generalized Normal Distribution} (\bgnd) to address both challenges. The classical Generalized Normal Distribution (GND) encompasses or closely approximates a wide range of parametric distributions commonly encountered in operational settings. Our proposed \bgnds employs an ML method called gradient boosting \citep{friedman2001greedy} to estimate the location and scale parameters of the GND as flexible functions of covariates using regression tree learners. Thus \bgnds combines the flexibility of ML with operations domain knowledge. 
Besides being the first time that the GND has been studied from an ML perspective, we also provide the first proof for \bgnd's statistical consistency. This result automatically extends to all special cases of the \bgnds studied in the ML literature \citep{ngboost, marz2022distributional}, for which no consistency results have been established. 

Our motivating application comes from forecasting patient wait and service times in emergency departments (ED), an important question in healthcare operations \citep{shi2016models, song2015diseconomies, chan2017impact, niewoehner2023physician, chen2023using}. In particular, \cite{arora2023probabilistic} pioneers the case for distributional forecasts over point predictions, and is the only study to provide one for wait times. They use Quantile Regression Forests (QRF) \citep{meinshausen2006quantile} to obtain the forecasts, which is a popular nonparametric distributional ML technique. By leveraging operations knowledge, we show that \bgnds outperforms QRF on data from a large academic ED in the United States. Specifically, using the exponential case of \bgnds for wait times and the log-normal case for service times, \bgnds outperforms by 6.1\% and 8.8\% on the Continuous Ranked Probability Score (CRPS) for out-of-sample wait and service time forecasts, respectively (see Table~\ref{tab:crps}). Further analysis shows that these improvements translate into a 9.4\% increase in patient satisfaction and a 4.1\% reduction in mortality for myocardial infarction patients, respectively (see \ref{sec:experiments} in the electronic companion
for details).

Perhaps more fundamentally, our approach highlights the value of incorporating operations domain knowledge: 
Using classic statistical models without any ML, the linearly-specified exponential and log-normal models already outperform QRF by 2.5\% and 7.0\% respectively in CRPS accuracy for patient wait and service times forecasts (Table~\ref{tab:crps}).
To our knowledge, this paper is the first to articulate and show that simple models equipped with operations knowledge can outperform nonparametric ML methods. The fact that \bgnds outperforms both the classical parametric models as well as QRF further demonstrates the value of combining ML techniques with operations domain knowledge in distributional forecasts.

\begin{table}
\caption{\label{tab:crps}\%Reduction in wait and service time CRPS values relative to QRF}
\centering
\begin{tabular}[t]{lcccclccc}
\toprule
\multicolumn{1}{c}{} & \multicolumn{4}{c}{Wait times} & \multicolumn{4}{c}{Service times} \\
\cmidrule(l{3pt}r{3pt}){2-5} \cmidrule(l{3pt}r{3pt}){6-9}
 & 2017 & 2018 & 2019 & Aggregate & 2017 & 2018 & 2019 & Aggregate\\
\midrule
\bgnd & 4.3\% & 8.3\% & 5.7\% & 6.1\% & 6.9\% & 7.4\% & 13\% & 8.8\%\\
Classic exponential/log-normal & 0.95\% & 5.1\% & 1.3\% & 2.5\% & 5.7\% & 6.7\% & 9.0\% & 7.0\%\\
\bottomrule
\end{tabular}
\end{table}

In terms of methodological contribution, while procedures exist for fitting special cases of the \bgnds such as the boosted normal distribution \citep{ngboost,marz2022distributional}, their statistical consistency have not been studied. A possible reason is that these procedures work by jointly minimizing the negative log-likelihood (the loss) over both the location and scale parameters via gradient descent. However, the population version of the loss is non-convex even for the normal distribution (see \ref{sec:non-convex} in the electronic companion for a numerical example). In addition, these procedures encounter computational issues when the log-likelihood is non-differentiable, as is the case with the Laplace distribution, which is a special case of GND. In such scenarios, the likelihood is not smooth in the location parameter, leading to numerical convergence failures. 

In contrast, we employ a natural insight from traditional statistics that decouples the estimation into two separate smooth convex problems: First we estimate the location parameter as the conditional mean, and then plug in that estimate to estimate the scale parameter. This mirrors how the mean and standard deviation are traditionally estimated. Moreover, this decoupling allows us to establish the consistency of \bgnd, and hence the boosted normal and Laplace distributions as well. In this sense, our approach adds both computational and theoretical value to the literature.

Finally, we also contribute to the growing literature that integrates operations concepts into machine learning. Recent studies propose data-driven optimization frameworks that consider the structure of the final optimization problem during parameter estimation. Some approaches are applicable to general contexts, such as problems with linear objective and constraints \citep{bertsimasPredictive2020, elmachtoubSmart2022}, while others focus on specific problems like the newsvendor problem \citep{banBig2019, notzPrescriptive2022, alleyPricing2023, singhFeatureDriven}. This research stream leverages knowledge of the optimization problem's structure, whereas our approach leverages distributional knowledge.

\section{Boosted GND}

The classic GND has location and scale parameters $\mu\in\Real$ and $b>0$ respectively, and probability density 
$$
p(y;\mu,b,\gamma) = \left\{2\gamma^{1/\gamma}\Gamma(1+1/\gamma)\right\}^{-1}  b^{-1}\exp\left(-\frac{|y-\mu|^{\gamma}}{\gamma b^\gamma}\right),
$$
where different values of the shape parameter $\gamma\ge 1$ yields various distributions characterized in the literature for operations problems. When $\gamma=1$ we recover the Laplace distribution. For $\gamma=2$ we have the normal distribution, with $\mu$ representing the mean and $b$ the standard deviation. Moreover, if $V$ follows a log-normal distribution (e.g. service times), then $\log V$ is normal, and $\Prob(V\le v) = \Phi(\frac{\log v - \mu}{b})$ for some $\mu$ and $b$.
Similarly, if $V$ follows an exponential distribution (e.g. wait times), then $V^{1/4}$ is very near normality.\footnote{If $V\sim Exp(r)$, then $2r V \sim \chi_2^2$, and the normal approximation follows from the chi-squared approximation in \citet{hawkins1986note}.}
More generally, other types of distributions can also be well approximated using the Box-Cox family of transformations to normality~\citep{boxcox}, which includes the last two examples as special cases.

In classical parametric modelling, special cases of the GND allow for $\mu$ and/or $b$ to be heterogeneous in covariates $x\in\X\subset\Real^p$. For example the standard log-normal regression model is $ \log V = x'\beta + \e$, where $\e \sim N(0,\sigma^2)$.\footnote{Throughout the article, the prime notation in the dot product $v^\prime w$ indicates that $v^\prime$ is the transpose of the vector $v$.} In this case the shape parameter is fixed at $\gamma=2$ and the location and scale parameters are specified linearly as $\mu(x) = x'\beta, b(x) = \sigma$.
The shape parameter $\gamma$ is typically fixed for all $x$. For example, it would be unusual for the distribution to be log-normal for some values of $x$, but change shape completely and become log-Laplace for other values of $x$.

Being able to relax the linear specifications above to allow $\mu(x)$ and $b(x)$ to be flexible functions of $x$ may yield better fit to the data, while still maintaining the parametric GND form for statistical efficiency. This paper proposes a novel distributional ML approach for achieving this. Specifically, we propose a consistent ML estimator for the parameters of the GND. The boosted GND has density
\begin{equation}\label{eq:boostedGND}
\mathit{bGND}(\mu,b) \sim b(x)^{-1}\exp\left(-\frac{|y-\mu(x)|^{\gamma}}{\gamma b(x)^\gamma}\right).
\end{equation}
Here, we estimate the true parameter functions $\mu(x)$ and $b(x)$ with ensembles of regression tree learners $\muhat(x)$ and $\bhat(x)$, fit using a ML method based on gradient boosting \citep{friedman2001greedy} that is described in Section~\ref{sec:algo_intro}. The value of $\gamma$ is set exogeneously based on domain knowledge, and is assumed not to vary with $x$ as previously discussed. Thus~\eqref{eq:boostedGND} provides a unifying model for analyzing a number of parametric distributions in one go. Importantly, we establish the statistical consistency of \bgnds in Section~\ref{sec:theory}.

We first illustrate the importance of leveraging operations knowledge by demonstrating how classical parametric models informed by the operations literature, can outperform distribution-agnostic ML approaches. We then present \bgnds and study its statistical properties.

\subsection{The Value of Operations-Informed Parametric Models}\label{subsec:ops_value}

We highlight the value of operations knowledge by examining the challenges many EDs face in forecasting patient wait and service times. The former is the time between ED entry and assignment to an ED bed, and the latter is the time between bed assignment and discharge/inpatient admission. We have data on patient visits to a large academic ED in the United States between 2016 and 2019 inclusive. Section \ref{sec:results} provides more details on the dataset, which includes patient-level covariates on demographics, medical information such as complaint category, and sojourn timestamps. Considering the size and dimensionality of the data, it seems appropriate to follow \citet{arora2023probabilistic} in applying QRF to forecast individualized wait time distributions for patients. QRF is a random forests algorithm that nonparametrically estimates distributions conditional on covariates, meaning that it does not assume that the data come from any particular class of distributions. In contrast, as discussed in the Introduction, the operations literature has shown that wait and service times are well approximated by exponential and log-normal distributions respectively. Thus, an alternative approach to QRF is to simply use the classic exponential and log-normal models whose parameters are specified linearly in the covariates.

\begin{figure}[h!]
    \centering
    \caption{\label{fig:waittime_hist_train} Patient wait time density conditional on time of arrival}
    \includegraphics[width=1\textwidth]{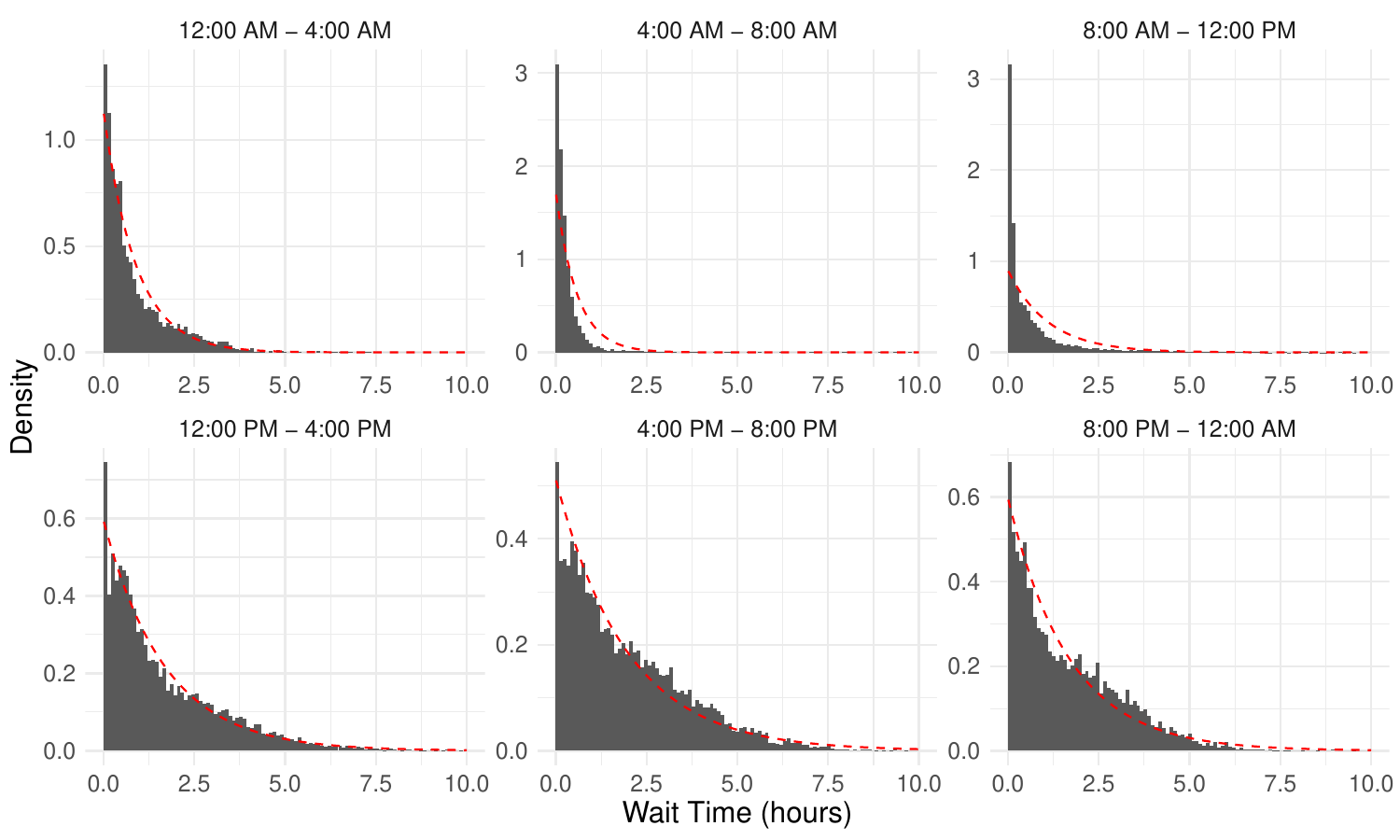}
        \begin{flushleft}
    \small
    \textit{Note: } The histograms depict the empirical density of patient wait times within each arrival time interval. The red dashed lines are the densities of the fitted exponential distribution conditional on being in each interval. The model we propose will systematically forecast the distributions by conditioning on a richer set of features described in Section~\ref{sec:results}.
    \end{flushleft}
    
\end{figure}

Figure~\ref{fig:waittime_hist_train} displays the histograms of wait times in our data. The histograms are bucketed by time of arrival, the predictor that explains the most variation in wait times in our data. Essentially, the histograms provide a discrete approximation to the conditional distribution of wait times given an arrival time interval. Within each interval, an exponential distribution is fit to the data (red dotted line). As predicted by the operations literature, the fitted exponential distributions agree remarkably well with the associated empirical histograms, especially considering that we only use time of arrival as the sole predictor.

Given this, we test the out-of-sample predictive performance of the classic exponential model where the log-rate parameter $\log\lambda$ is linear in the covariates $x$, i.e. $\lambda = e^{x^\prime\beta}$.
We fit this model as well as QRF to 2016 data, and then use both models to forecast wait times in the 2017 test period. We repeat the process for the 2018 and 2019 test periods using the preceding year as the training period. As seen in Table~\ref{tab:crps}, the classic exponential model actually results in a CRPS\footnote{CRPS is commonly used to evaluate probabilistic forecasts, see~\citet{gneiting2007strictly}. \cite{arora2023probabilistic} also use it to quantify the accuracy of ED wait time distributional forecasts. The CRPS generalizes the mean squared error for point predictions to distributional forecasts, hence a lower value implies higher accuracy.} value that is 2.5\% lower than that for QRF (lower values are better). In other words, the operations-informed parametric model outperforms the distribution-agnostic ML algorithm in this setting.

Similarly, the histograms for service times are consistent with a log-normal distribution (see Figure \ref{fig:servicetime_hist_train} the electronic companion). We thus compare the out-of-sample predictive performance of the classic log-normal distribution to the QRF. Specifically, we specify the parameters of the log-normal distribution linearly in $x$, i.e. $\mu_{LN} = x^\prime\gamma$ and $\log\sigma_{LN} = x^\prime\theta$, and estimate $(\gamma,\theta)$ by maximum likelihood. Once again we find that the classic log-normal model has a CRPS value that is 7.0\% lower than that for QRF.

Overall, our results demonstrate that the operations-informed parametric models outperform the nonparametric machine learning QRF in both settings. This underscores the value of operations domain knowledge, demonstrating that it can surpass complex and flexible ML techniques. \bgnd, which we describe next, is designed to further enhance performance by integrating this knowledge into distributional machine learning.

\subsection{Estimation Algorithm for \bgnd}\label{sec:algo_intro}




The approach we propose for estimating \bgnd~\eqref{eq:boostedGND} is based on the simple observation that there is no need to jointly estimate $\mu(x)$ and $b(x)$, which is what existing approaches do for special cases of \bgnds such as boosted normal distribution. The location parameter $\mu(x)$, being equal to the conditional mean $\E(Y|x)$ in a GND, can first be estimated using any boosted least-squares regression method that is consistent. We then show that the scale parameter $b(x)$ can be consistently estimated by setting $\mu(x)$ equal to the estimated $\hat\mu(x)$. Our approach decouples the estimation into two smooth convex problems even when $\gamma=1$ (i.e. the Laplace distribution), thereby sidestepping the numerical issues faced by existing approaches. Our approach is also more natural as it mirrors how the mean and variance are traditionally estimated.

We first reparameterize the scale parameter as
\begin{equation}\label{eq:scale_param}
b(x)=e^{-\beta(x)/\gamma}
\end{equation}
where $\beta(x)\in\Real$. The log-density of~\eqref{eq:boostedGND} then becomes $\beta(x)-|y-\mu(x)|^{\gamma}e^{\beta(x)}$ up to a multiplicative factor. Thus given a sample $\samp=\{(x_{i},y_{i})\}_{i=1}^{n}$ of i.i.d. copies of $(X,Y)$ where $Y|X \sim GND(\mu(X),b(X))$, the scaled negative log-likelihood is
\begin{equation}\label{eq:emp_risk}
R_{n}(\mu,\beta,\samp)= |\samp|^{-1} \sum_{i\in\samp}\left\{ |y_{i}-\mu(x_{i})|^{\gamma}e^{\beta(x_{i})}-\beta(x_{i})\right\}.
\end{equation}
We will refer to~\eqref{eq:emp_risk} as the likelihood risk, and its population version is
\begin{equation}\label{eq:pop_risk}
R(\mu,\beta) = \E_{X,Y} R_n(\mu,\beta, \{(X,Y)\}) = \int \left\{ \E_{Y|x}( |Y-\mu(x)|^\gamma ) e^{\beta(x)} - \beta(x)
\right\} dP_X,
\end{equation}
where $P_X$ is the distribution of $X\in\X\subset\Real^p$. Its empirical analogue is $P_{X,n_2}(B) = n_2^{-1}\sum_{i\in\samp_2} I(x_i \in B)$.

\begin{algorithm}[htpb]
\centering
\caption{Fitting the \bgnds~\eqref{eq:boostedGND} \label{alg:bgnd}}
\begin{algorithmic}[1]
  \STATE Split $\samp = \{ (x_i,y_i) \}_{i=1}^n$ into two random subsamples $\samp_1$ and $\samp_2$ of equal size, $n_1 = n_2 = n/2$
  \STATE Use any tree-boosted regression method satisfying~\eqref{eq:muhat_property} to estimate $\mu(x)$ from $\samp_1$. Call this $\muhat(x)$
  \STATE Apply a tree-boosted algorithm such as Algorithm~\ref{alg:genericboost} in Appendix~\ref{sec:alg2} to minimize the empirical risk $R_n(\muhat,\beta,\samp_2)$ over $\samp_2$. The resulting tree-boosted minimizer is $\betahat(x)$
  \STATE The tree-boosted estimators for the location and scale parameters are
  $$
  \muhat(x), \quad \hat b(x) = e^{-\betahat(x)/\gamma}
  $$
\end{algorithmic}
\end{algorithm}

Algorithm~\ref{alg:bgnd} describes our approach for estimating $\mu(x)$ and $\beta(x)$. The analysis of the algorithm is facilitated by sample splitting, an idea from the statistics literature that goes back to at least \citet{bickel1982}. First, we randomly split the sample $\samp$ into two disjoint subsamples $\samp_1$ and $\samp_2$ of equal size $n_1 = n_2 = n/2$.\footnote{We can improve efficiency by reversing the roles of $\samp_1$ and $\samp_2$, which yields another pair of consistent estimators $(\muhat'(x),\betahat'(x))$. The average $\frac{1}{2}(\muhat(x)+\muhat'(x),\betahat(x)+\betahat'(x))$ is more efficient than either one alone, and this approach will be used in our empirical analyses. However, the question of maximal efficiency is beyond the scope of this paper, and we defer this to future research.} Next, noting that $\mu(x)$ is the conditional mean $\E(Y|x)$, employ any consistent tree-boosted least-squares regression method (for example, \citet{zhangyu}) to obtain an estimate $\muhat(x)$ from $\samp_1$ that satisfies\footnote{Consistency results for boosted regression are typically in terms of pointwise convergence~\citep{zhangyu}. However, under the assumptions in Section~\ref{sec:theory}, pointwise convergence implies uniform convergence~\eqref{eq:muhat_property}: Under Assumption~\ref{asm:true_parm} it suffices for $\muhat$ to converge uniformly to $\mustar$ in~\eqref{eq:parm_star}, and under Assumption~\ref{asm:x_bound} both functions are piecewise-constant over a finite number of regions in $\X$. Thus it suffices for $\muhat$ to converge pointwise to $\mustar$.}
\begin{equation}\label{eq:muhat_property}
\|\mu-\muhat\|_\infty = \e_{n_1} \rightarrow_p 0 \text{ w.r.t. } \samp_1.
\end{equation}
The final step plugs $\muhat(x)$ into the empirical risk~\eqref{eq:emp_risk} evaluated on $\samp_2$, 
$R_n(\muhat,\beta,\samp_2)$, and employs a gradient-boosted algorithm to minimize $R_n(\muhat,\beta,\samp_2)$ over $\beta\in\Fspace$, where $\Fspace$ is the span of regression trees characterized in Section~\ref{sec:theory}. Algorithm~\ref{alg:genericboost} in Appendix~\ref{sec:alg2} provides one way to do this. The resulting tree-boosted minimizer is our estimate $\betahat(x)$, and our estimate for the scale parameter is $\bhat(x) = e^{-\betahat(x)/\gamma}$.

\section{Statistical Consistency of \bgnd}\label{sec:theory}
In Section~\ref{subsec:consistency} we show the statistical consistency of our approach, the paper's main theoretical result. Supporting results can be found in Section~\ref{sub:supportingresults}. Our analysis rests on four conditions. The first three are standard for the nonparametric estimation of functions, i.e. $\mu(x)$ and $b(x)$ in our context. 
\begin{assumption}\label{asm:x_bound}
The domain of the covariates $\X\subset\Real^p$ is bounded.
\end{assumption}
\begin{assumption}\label{asm:x_density}
The distribution $P_X$ of the covariates $X$ admits a density that is bounded away from zero and infinity on $\X$. Define also the empirical distribution $P_{X,n_2}(B) = n_2^{-1}\sum_{i\in\samp_2} I(x_i \in B)$.
\end{assumption}
\begin{assumption}\label{asm:parm_bound}
The true location parameter $\mu(x)$ is bounded between some interval
$[\mulo,\muhi] \subset (0,\infty)$ on $\X$, and the true scale parameter $b(x)$ is bounded between some interval
$[\blo,\bhi] \subset (0,\infty)$ on $\X$.
\end{assumption}

To motivate the final assumption, define the span of the regression tree functions that hosts the tree-boosted estimators $\muhat(x)$ and $\betahat(x)$ as
$$
\Fspace = \left\{ \sum_{k=1}^m g_k(x) : m\in\mathbb{N}, g_k(x) \text{ a regression tree function defined on } \X \right\},
$$
and for $\Psi>0$, its restriction
\begin{equation}\label{eq:Fspace_Psi}
\Fspace^{\Psi} = \left\{ F\in\Fspace : \| F \|_\infty < \Psi \right\}.
\end{equation}
In particular, $\Fspace^{\Psi_{n_2}}$ contains all the boosting iterates $b_0, b_1,\cdots,\bhat$ produced by Algorithm~\ref{alg:genericboost} in Appendix~\ref{sec:alg2}. Following \cite{lee2021boosted}, $\Fspace$ is equivalent to $\{ \sum_j c_j I_{B_j} : c_j \in \mathbb{R} \}$, i.e. the span of indicator functions over disjoint hypercubes of $\X$ indexed by $j=(j_1,\cdots,j_p)$:
\begin{equation}\label{eq:regions}
B_{j}=\left\{ \begin{array}{ccc}
x = (x^{(1)},\cdots,x^{(p)}) & : & \begin{array}{c}
x^{(1,j_{1})}<x^{(1)}\leq x^{(1,j_{1}+1)}\\
\vdots\\
x^{(p,j_{p})}<x^{(p)}\leq x^{(p,j_{p}+1)}
\end{array}\end{array}\right\}.
\end{equation}
The regions $B_j\subset\X$ are formed using all possible split points
$\{x^{(k,j_k)}\}_{j_k}$ for the $k$-th coordinate $x^{(k)}$, with the spacing determined by the precision of the measurements. For example, if weight is measured to the closest kilogram, then the set of all possible split points will be $\{0.5, 1.5, 2.5,\cdots\}$ kilograms. While abstract treatments of trees assume that there is a continuum of split points, in reality they fall on a discrete grid that is pre-determined by the precision of the data.

Given that $x$ is inevitably measured to somewhere within one of these regions,
what we can identify from data is not $\mu(x)$ or $b(x)^\gamma$, but the coarsened versions
\begin{equation}\label{eq:parm_star}
\begin{aligned}
\mu^\ast(x)\left|_{B_j}\right. &:= \E\{ \mu(X) | X\in B_j \} = P_X(B_j)^{-1}\int_{B_j} \mu(x)dP_X, \\
\{b^\ast(x)\}^\gamma\left|_{B_j}\right. &:= \E\{ b(X)^\gamma | X\in B_j \} = P_X(B_j)^{-1}\int_{B_j} b(x)^\gamma dP_X,
\end{aligned}
\end{equation}
i.e. the true $\mu(x)$ and $b(x)$ are identified up to a piecewise-constant approximation within each fundamental partition $B_j$, i.e. $\mu^\ast, b^\ast \in \Fspace$. Recall that this is a measurement precision issue, so it holds regardless of the algorithm we use to estimate the parameters. The approximation errors $\|\mu-\mu^\ast\|_\infty$ and $\|b-b^\ast\|_\infty$ are thus irreducible, and we consider them to be negligible in a numerical tolerance sense.
\begin{assumption}\label{asm:true_parm}
$\|\mu-\mu^\ast\|_\infty = \|b-b^\ast\|_\infty = 0$.
\end{assumption}
\begin{remark}
If $\mu(x)$ and $b(x)$ are sufficiently smooth, $\mu^\ast(x)$ and $b^\ast(x)$ will closely approximate them within a small enough partition $B_j$ even without Assumption~\ref{asm:true_parm}. To see this, suppose that $b(x)$ is H\"older continuous, i.e. $|b(x)-b(x')| \precsim \|x-x'\|^\alpha$ for some $\alpha>0$. Then $\inf_{x\in B_j} b(x) \le b^\ast(x)|_{B_j} \le \sup_{x\in B_j} b(x)$ and
$$
\left\| b - b^\ast \right\|_\infty \precsim \max_j(\text{diam}B_j)^\alpha.
$$
A similar result holds for $\left\| \mu - \mu^\ast \right\|_\infty$. When the covariates are measured to high precision, the diameter of the $B_j$'s will be small, which implies close uniform approximation of $b$ by $b^\ast$ and of $\mu$ by $\mu^\ast$.
\end{remark}
\begin{remark}\label{rem:minprob}
Assumption~\ref{asm:x_density} implies the following, which will be useful in various parts of the proof:
$$
\minprob = \min_j P_X(B_j) > 0.
$$
\end{remark}
\begin{remark}
Since $\Fspace$ is the linear span of piecewise-constant functions, it is closed under pointwise exponentiation, i.e. $f\in\Fspace \Rightarrow e^f\in\Fspace$. Likewise, $f\in\Fspace \Rightarrow f^{1/\gamma}\in\Fspace$. Hence $\mu^\ast, b^\ast, \bhat \in \Fspace$.
\end{remark}

\subsection{Consistency of $(\muhat,\bhat)$}\label{subsec:consistency}

We are now in a position to state our main result. For $Y|x \sim bGND(\mu(x),b(x))$, $\muhat$ is by default a consistent estimator for $\mu$ due to~\eqref{eq:muhat_property}, so it remains to show that the estimator $\bhat$ for the scale is also consistent. Proposition~\ref{prop:main} below can also be expressed in non-asymptotic terms but we omit the analysis for brevity.

\begin{proposition}\label{prop:main}
Under Assumptions~\ref{asm:x_bound}-\ref{asm:true_parm},
$$
\| \bhat - b \|_\infty = o_p(1).
$$
\end{proposition}

To establish the result, we first show that the minimizer $\betastar(x) = \arg\min_{\beta(x)\in\Fspace} R(\muhat,\beta)$ exists, and denote $\bstar(x) = e^{-\betastar(x)/\gamma}$. Given that $\Fspace$ is the span of piecewise-constant functions over the fundamental partitions $\{B_j\}_j$ in~\eqref{eq:regions}, we can write
$\muhat(x) = \sum_j \muhat|_{B_j}I_{B_j}(x)$, $\betastar(x) = \sum_j \betastar|_{B_j}I_{B_j}(x)$, and $\bstar(x) = \sum_j e^{-\betastar|_{B_j}/\gamma}I_{B_j}(x)$.

\begin{lemma}\label{lem:betastar}
If $\| \mu-\muhat \|_\infty = \e_{n_1} < \infty$ in~\eqref{eq:muhat_property} then $\betastar$ exists and its value inside $B_j$ satisfies
$$
\bstar(x)^\gamma|_{B_j} = e^{-\betastar|_{B_j}} = P_X(B_j)^{-1} \int_{B_j} \E_{Y|x} \left|Y-\muhat|_{B_j} \right|^\gamma dP_X = \E( \left|Y-\muhat|_{B_j} \right|^\gamma | X\in B_j ).
$$
Furthermore $\|\bstar-b\|_\infty < \e_{n_1}$, and lastly, if $\e_{n_1} \le \blo/2$ then
$$
\blo/2 \le \bstar(x) \le 2\bhi \text{ and } -\gamma\log(2\bhi) \le \betastar(x) \le \gamma\log(2/\blo).
$$
\end{lemma}

\proof{Proof.} 
Since $\|\muhat\|_\infty \le \e_{n_1} + \|\mu \|_\infty < \infty$,
$$
\begin{aligned}
R(\muhat,\beta) &= \E_{X,Y} R_n(\muhat,\beta, \{(X,Y)\}) = \sum_j \int_{B_j} \left\{ \E_{Y|x} \left|Y-\muhat|_{B_j} \right|^\gamma e^{\beta|_{B_j}} - \beta|_{B_j} \right\} dP_X \\
&= \sum_j \left\{ e^{\beta|_{B_j}} \int_{B_j} \E_{Y|x} \left|Y-\muhat|_{B_j} \right|^\gamma dP_X - \beta|_{B_j} P_X(B_j) \right\}.
\end{aligned}
$$
By Remark~\ref{rem:minprob}, $P_X(B_j)>0\,\, \forall j$ so the first order condition for each $\beta|_{B_j}$ implies the first claim. For the second claim,
$$
\begin{aligned}
\bstar\left|_{B_j}\right. = e^{-\betastar|_{B_j}/\gamma}
&= \left( \int_{B_j} \E_{Y|x}\left|Y-\muhat|_{B_j} \right|^{\gamma} d\frac{P_X}{P_X(B_j)} \right)^{1/\gamma} = \left\| Y-\muhat \right\|_{B_j,\gamma} \\
&\le \left\| Y-\mu \right\|_{B_j,\gamma} + \left\| \mu - \muhat \right\|_{B_j,\gamma} \le \left\| Y-\mu \right\|_{B_j,\gamma} + \e_{n_1} = b^\ast|_{B_j} + \e_{n_1},
\end{aligned}
$$
where the last equality comes from noting that $\E_{Y|x}\left|Y-\mu(x)\right|^{\gamma} = b(x)^\gamma$ and from the definition of $b^\ast(x)$ in~\eqref{eq:parm_star}. Reversing the roles of $\bstar$ and $b^\ast$ shows that $\| \bstar-b \|_\infty \le \|\bstar - b^\ast\|_\infty + \|b^\ast - b\|_\infty < \e_{n_1}$ by Assumption~\ref{asm:true_parm}.
For the final claim, note from~\eqref{eq:parm_star} that $\blo \le b^\ast(x) \le \bhi$. Since $\|\bstar - b^\ast\|_\infty \le \e_{n_1}$ and $\e_{n_1} \le \blo/2$ by hypothesis,
$$
\blo/2 = \blo -\blo/2 \le b^\ast(x) - \e_{n_1} \le \bstar(x) \le b^\ast(x) + \e_{n_1} \le \bhi + \bhi = 2\bhi,
$$
and the bound for $\betastar$ follows immediately from $\bstar(x) = e^{-\betastar/\gamma}$.
\hfill\Halmos
\endproof

\proof{Proof of Proposition~\ref{prop:main}.}

From~\eqref{eq:muhat_property} and Lemma~\ref{lem:betastar} we have $\|\hat b-b\|_\infty \le \|\hat b-\bstar\|_\infty + \e_{n_1}$ where $\e_{n_1}=o_p(1)$ w.r.t. the sample $\samp_1$, so eventually $\e_{n_1} \le \blo/2$ with high probability, thus satisfying the hypothesis in Lemma~\ref{lem:betastar}. It then suffices to show that $\|\hat b-\bstar\|_\infty = o_p(1)$ w.r.t. the sample $\samp_2$. By virtue of $\betastar$ being the minimizer of the population risk $R(\muhat,\beta)$ in~\eqref{eq:pop_risk}, there exists $\rho\in(0,1)$ for which Taylor's theorem yields the first equality below:
$$
\begin{aligned}
0   &\le R(\muhat,\betahat) - R(\muhat,\betastar) = \frac{1}{2} \int \E_{Y|x}\{ |Y-\muhat(x)|^\gamma \} e^{ \betastar(x)+\rho\{\betahat(x)-\betastar(x)\}}
    \cdot \{\betahat(x)-\betastar(x)\}^2 dP_X \\
    &= \frac{1}{2} \sum_j e^{ \rho(\betahat|_{B_j}-\betastar|_{B_j}) }
    \cdot (\betahat|_{B_j}-\betastar|_{B_j})^2  P_X(B_j) \ge \frac{1}{2} e^{-\|\betahat-\betastar\|_\infty} \minprob \| \betahat - \betastar \|_\infty^2,
\end{aligned}
$$
where the second equality follows from noting that all terms apart from $\E_{Y|x}\{ |Y-\muhat(x)|^\gamma \}$ is constant within each region $B_j$, and that $\int_{B_j} \E_{Y|x} \left|Y-\muhat|_{B_j} \right|^\gamma dP_X = e^{-\betastar|_{B_j}} P_X(B_j)$ from Lemma~\ref{lem:betastar}. Suppressing the notation $\samp_2$ in $R_n(\muhat,\beta,\samp_2)$, it follows from the inequality $|e^u-e^v|\le\max(e^u,e^v)|u-v|$ that
\begin{equation}\label{eq:ERM}
\begin{aligned}
\| \hat b - \bstar \|_\infty
&\le
 \max_{x\in\X}\{\hat b(x) \vee \bstar(x) \} \| \betahat - \betastar \|_\infty/\gamma
 \le \underbrace{ \frac{ \max_{x\in\X}\{\hat b(x) \vee \bstar \}
e^{\frac{1}{2}\|\betahat-\betastar\|_\infty} }{ \gamma (\minprob/2)^{1/2} } }_{C}
\{ R(\muhat,\betahat) - R(\muhat,\betastar) \}^{1/2} \\
&\le C\{
\underbrace{R(\muhat,\betahat) - R_n(\muhat,\betahat)}_{(i)}
+
\underbrace{R_n(\muhat,\betahat) - R_n(\muhat,\betastar)}_{(ii)}
+
\underbrace{R_n(\muhat,\betastar) - R(\muhat,\betastar)}_{(iii)} \}^{1/2}.
\end{aligned}
\end{equation}
The fraction $C$ is bounded with high probability w.r.t. the sample $\samp_2$ because $\min_j P_X(B_j)>0$ by Remark~\ref{rem:minprob}, $\{ \bstar(x), \betastar(x) \}$ are bounded by Lemma~\ref{lem:betastar}, and $\{ \betahat(x), \bhat(x) \}$ are bounded with high probability by Lemma~\ref{lem:bhat_bound}. The quantities (i) and (iii) are the deviations of the empirical risk from the population risk. Proposition~\ref{prop:main_conc} in Section~\ref{subsec:conc} develops concentration inequalities to show that both quantities are $o_p(1)$ w.r.t. $\samp_2$, bearing in mind that $\betahat\in\Fspace^{\Psi_{n_2}}$ \eqref{eq:Fspace_Psi}. Finally, (ii) represents the minimization of the empirical risk~\eqref{eq:emp_risk}. Proposition~\ref{prop:minrisk} in Section~\ref{subsec:minrisk} shows that this term is also $o_p(1)$ w.r.t. $\samp_2$.
\hfill\Halmos
\endproof

\subsection{Supporting results}\label{sub:supportingresults}

From now on we will suppress the notation $\samp_2$ from the empirical risk $R_n(\muhat,\beta,\samp_2)$. Let us also recall basic concepts for empirical processes from Chapter 2 of~\citet{vaart} that will be used in~$\S$\ref{subsec:conc}. For a probability measure $P$ on $\X$, the $L^{2}(P)$-ball of radius $\delta>0$
centred at some $\beta\in L^{2}(P)$ is
$\{\beta'\in\Fspace^{\Psi}: \int (\beta'-\beta)^2 dP <\delta^2 \}$ where $\Fspace^\Psi$ is defined in~\eqref{eq:Fspace_Psi}. The covering
number $\mathcal{N}(\delta,\Fspace^{\Psi},P)$ is the minimum number of such balls needed to cover $\Fspace^{\Psi}$, so $\mathcal{N}(\delta,\Fspace^{\Psi},P)=1$ for
$\delta\ge\Psi$. The entropy integral
\begin{equation}\label{eq:entintegral}
  J_{\Fspace}=\sup_{\Psi>0,P}
  \int_0^1 \sqrt{ \log\mathcal{N}(u\Psi,\Fspace^{\Psi}, P) } du
\end{equation}
represents the complexity of the function class $\Fspace$, and is at most finite because $\Fspace$ is the span of indicator functions over a finite number of regions~\eqref{eq:regions}. For convenience we will assume that $J_\Fspace>1$.

\subsubsection{Concentration inequalities.}\label{subsec:conc}

The results herein are greatly facilitated by decomposing $(X,Y)$ into independent components $X$ and $W$: Let $\{ W_i \}_{i\in\samp_2}$ be i.i.d. $GND(0,1)$ random variables that are independent of $\{X_i,Y_i\}_{i\in\samp_2}$. Then conditional on $X_i$, $Y_i' = \mu(X_i)+b(X_i)W_i$ has the same distribution as $Y_i|X_i$. Thus the empirical risk $R_n(\muhat,\beta)$ has the same distribution as
\begin{equation}\label{eq:alt_Rn}
R_n'(\muhat,\beta) = n_2^{-1}\sum_{i\in\samp_2} \left\{ |Y_i'-\muhat(X_i)|^{\gamma}e^{\beta(X_i)}-\beta(X_i) \right\}.
\end{equation}
The decomposition allows us to condition on the set $\mathcal{W} = \{|W_1|,\cdots,|W_{n_2}| \le 2\gamma\log n_2 \}$, on which
\begin{equation}\label{eq:clip1}
\max_{i\in\samp_2} |Y'_i-\muhat(X_i)| \le \bhi \max_{i\in\samp_2} |W_i| + \e_{n_1} < 3\gamma\bhi\log n_2
\end{equation}
when $\| \mu-\muhat \|_\infty = \e_{n_1}$ in~\eqref{eq:muhat_property} is less than $\blo/2$ eventually. Writing $Q_{n_2} = \Prob (|W|> 2\gamma\log n_2)$, the conditional expectation of an integrable $f$ on $\mathcal{X}^{n_2}\times\mathcal{W}$
can be re-expressed as an unconditional one:
\begin{equation}\label{eq:cond_exp}
\begin{aligned}
\E \{f(X,W)|\mathcal{W}\}
&= \E_{X_1,\cdots,X_{n_2}} \int_{|W_1|,\cdots,|W_{n_2}| \le 2\gamma\log n_2} f\cdot \prod_{i\in\samp_2}\frac{dP_{W_i}}{1-Q_{n_2}} \\
&= \E_{X_1,\cdots,X_{n_2}} \int f \cdot \prod_{i\in\samp_2} dP_{Z_i} = \E f(X,Z)
\end{aligned}
\end{equation}
where the i.i.d. $\{Z_i\}_{i\in\samp_2}$ are each bounded between $\pm 2\gamma\log n_2$, and are independent of $\{X_i,Y_i\}_{i\in\samp_2}$.

\begin{proposition}\label{prop:main_conc}
Recall that $P_{X,n_2}$ is the empirical measure defined in Assumption~\ref{asm:x_density}, and $\Fspace^\Psi$ is defined in~\eqref{eq:Fspace_Psi}. For $n_2 \ge 3$ and $ \| \mu-\muhat \|_\infty = \e_{n_1} \le \blo/2$ in~\eqref{eq:muhat_property}, there exist constants $\kappa_1,\kappa_2$ depending on $\gamma$, $\blo$, and $\bhi$ such that with probability $1 - \kappa_1 \exp\left\{ - \kappa_2 (\eta\wedge\minprob /J_\Fspace)^2 n_2^{2/5} \right\} - \gamma n_2^{-1}$, we have for $\eta\le 1$: 
$$
\min_j P_{X,n_2}(B_j) \ge \minprob/2,
\qquad
\sup_{\beta\in\Fspace^{\Psi_{n_2}}} \big| R_n(\muhat,\beta) - R(\muhat,\beta) \big| \le 2\eta, 
\qquad
\big| R_n(\muhat,\betastar) - R(\muhat,\betastar) \big| \le \eta,
$$
$$
\big| R_n(\muhat,0) - R(\muhat,0) \big| = \left| n_2^{-1}\sum_{i\in\samp_2} |Y_i-\muhat(X_i)|^\gamma - \E |Y_i-\muhat(X_i)|^\gamma \right| \le \eta.
$$
\end{proposition}
\proof{Proof.}
Note that $|Y'_i-\muhat(X_i)|^\gamma$ has a subexponential tail (consider for example the square of the normal distribution, i.e. chi-squared distribution). We therefore truncate it to facilitate a maximal inequality for the deviation of $R_n'(\muhat,\beta)$ in~\eqref{eq:alt_Rn} from $R(\muhat,\beta)$ based on subgaussian increments: By the hypotheses in the proposition, $\max_{i\in\samp_2}|Y'_i-\muhat(X_i)|^\gamma$ is bounded on $\mathcal{W}$ per~\eqref{eq:clip1}. Denoting the events
$$
U_{\Fspace^{\Psi_{n_2}}} = \left\{ \sup_{\beta\in\Fspace^{\Psi_{n_2}}} \big| \{R_n'(\muhat,\beta)-R_n'(\muhat,0)\} - \{R(\muhat,\beta)-R(\muhat,0)\} \big| > \eta \right\},
$$
$$
U_0 = \left\{ \big| R_n'(\muhat,0) - R(\muhat,0) \big| > \eta \right\},
\qquad
U_{\betastar} = \left\{ \big| R_n'(\muhat,\betastar) - R(\muhat,\betastar) \big| > \eta \right\},
$$
we seek a bound for
\begin{equation}\label{eq:clip2}
\begin{aligned}
\Prob\left( U_{\Fspace^{\Psi_{n_2}}} \cup U_0 \cup U_{\betastar} \right)
&\le \Prob\left( U_{\Fspace^{\Psi_{n_2}}} \cup U_0 \cup U_{\betastar}, \mathcal{W}\right) + \Prob( \mathcal{W}^c )
\le \Prob\left(\left. U_{\Fspace^{\Psi_{n_2}}} \cup U_0 \cup U_{\betastar} \right| \mathcal{W} \right) + \gamma n_2^{-1} \\
&\le \Prob\left(\left. U_{\Fspace^{\Psi_{n_2}}} \right| \mathcal{W} \right) + \Prob\left(\left. U_0 \right| \mathcal{W} \right) + \Prob\left(\left. U_{\betastar} \right| \mathcal{W} \right) + \gamma n_2^{-1},
\end{aligned}
\end{equation}
where the second inequality comes from Lemma~\ref{lem:gnd_tailbound} and the fact that $P(A\cap B) \le P(A|B)$. An issue with bounding these probabilities is that, by conditioning on $\mathcal{W}$, $R_n'(\muhat,\beta)$ concentrates around $\E\{R_n'(\muhat,\beta)|\mathcal{W}\}$ rather than $R(\muhat,\beta)$. Thus define instead
$$
V_\beta = \{R_n'(\muhat,\beta)-R_n'(\muhat,0)\} - [ \E\{R_n'(\muhat,\beta)|\mathcal{W} \} - \E\{ R_n'(\muhat,0)|\mathcal{W} \} ],
$$
$$
V_0 = R_n'(\muhat,0) - \E\{ R_n'(\muhat,0)|\mathcal{W} \},
\qquad
V_{\betastar} = R_n'(\muhat,\betastar) - \E\{ R_n'(\muhat,\betastar)|\mathcal{W} \}.
$$
Lemma~\ref{lem:cond_conc} shows that the probabilities in~\eqref{eq:clip2} can be bounded in terms of the $V$'s. This is used in Lemmas~\ref{lem:ulln} and \ref{lem:lln}, which collectively imply
$$
\begin{aligned}
&
\Prob\left( U_{\Fspace^{\Psi_{n_2}}} \cup U_0 \cup U_{\betastar} \right)
\le
\kappa_3 \exp\left\{ -\kappa_4 (\eta/J_\Fspace)^2 n_2^{2/5} \right\} + 2\kappa_5 \exp\left( -\kappa_6 \eta^2 n_2^{1/2} \right) + \gamma n_2^{-1} \\
&\qquad\le
(\kappa_3 + 2\kappa_5) \exp\left\{ -(\kappa_4\wedge\kappa_6) (\eta/J_\Fspace)^2 n_2^{2/5} \right\} + \gamma n_2^{-1}
=
\kappa_1 \exp\left\{ -\kappa_2 (\eta/J_\Fspace)^2 n_2^{2/5} \right\} + \gamma n_2^{-1},
\end{aligned}
$$
where the second inequality follows from assuming that $J_\Fspace > 1$ in~\eqref{eq:entintegral}. Note from Lemma~\ref{lem:ulln} that $\min_j P_{X,n_2}(B_j) \ge \minprob/2$ with probability $1-\kappa_3\exp\{ -\kappa_4 ( \minprob/J_\Fspace )^2 n_2 \}$, so by enlarging $\kappa_1$ we have both events $\left\{ \min_j P_{X,n_2}(B_j) \ge \minprob/2 \right\}$ and $\left\{ U_{\Fspace^{\Psi_{n_2}}}^C \cap U_0^C \cap U_{\betastar}^C \right\}$ hold with probability $1 - \kappa_1 \exp\left\{ -\kappa_2 (\eta\wedge\minprob /J_\Fspace)^2 n_2^{2/5} \right\} - \gamma n_2^{-1}$. Finally, noting that $\big| R_n(\muhat,\beta) - R(\muhat,\beta) \big| \le 2\eta$ on $U_{\Fspace^{\Psi_{n_2}}}^C \cap U_0^C$ concludes the proof.
\hfill\Halmos
\endproof

\begin{lemma}\label{lem:ulln}
There exist constants $\kappa_3$, $\kappa_4$ depending on $\gamma$, $\blo$, and $\bhi$ for which $\min_j P_{X,n_2}(B_j) \ge \minprob/2$ with probability $1-\kappa_3\exp\{ -\kappa_4 ( \minprob/J_\Fspace )^2 n_2 \}$, and 
$$
\Prob\left(\left. U_{{\Fspace^{\Psi_{n_2}}}} \right| \mathcal{W} \right) \le \kappa_3 \exp\left\{ -\kappa_4 (\eta/J_\Fspace)^2 n_2^{2/5} \right\}.
$$
\end{lemma}

\proof{Proof.}
From Lemma~\ref{lem:cond_conc} we have
$\Prob\left(\left. U_{{\Fspace^{\Psi_{n_2}}}} \right| \mathcal{W} \right)
\le \Prob( \sup_{\beta\in\Fspace^{\Psi_{n_2}}} |V_\beta| >\eta-\tau_1/n_2^{3/2} | \mathcal{W} )$. To bound this, define the Orlicz norm $\|X\|_{\Phi}=\inf\{C>0:\E\Phi(|X|/C)\le 1\}$ where
$\Phi(x)=e^{x^2}-1$. Taking expectations w.r.t. $(X,Z)$ in~\eqref{eq:cond_exp}, suppose the following bounds hold for some constant $\tau_2$ depending on $\gamma$, $\blo$, and $\bhi$:
\begin{equation}\label{eq:esup}
\left\| \sup_{\beta\in\Fspace^{\Psi_{n_2}}} |V_\beta| \right\|_{\Phi}
\le \tau_2 J_{\Fspace}/n_2^{1/5},
\qquad
\left\| \sup_{\beta\in\Fspace^1} \left| n_2^{-1}\sum_{i\in\samp_2}\beta(X_i) - \E_X\beta(X) \right| \right\|_{\Phi}
\le \tau_2 J_{\Fspace}/n_2^{1/2}.
\end{equation}
Applying Markov's inequality to the first inequality in~\eqref{eq:esup} implies that for $\eta > \tau_1/n_2^{3/2}$,
$$
\begin{aligned}
&
\quad \Prob\left( \left. \sup_{\beta\in\Fspace^{\Psi_{n_2}}} |V_\beta| >\eta-\tau_1/n_2^{3/2} \right| \mathcal{W} \right) \\
\le &
2\exp\left\{ -\left( \frac{\eta-\tau_1/n_2^{3/2}}{\tau_2 J_\Fspace} \right)^2 n_2^{2/5} \right\}
\le
2\exp\left\{ \frac{2\tau_1\eta}{(\tau_2 J_\Fspace)^2 n_2^{11/10}}
- \left( \frac{\eta}{\tau_2 J_\Fspace} \right)^2 n_2^{2/5} \right\} \\
\le &
2\exp\left\{ \frac{2\tau_1}{\tau_2^2 3^{11/10}} - \tau_2' (\eta/J_\Fspace)^2 n_2^{2/5} \right\}
=
\kappa_3 \exp\left\{ -\kappa_4 (\eta/J_\Fspace)^2 n_2^{2/5} \right\},
\end{aligned}
$$
recalling that $J_\Fspace>1$ in the last inequality. Noting that the indicators $I_{B_j}(\cdot)$ over the regions~\eqref{eq:regions} belong in $\Fspace^1$, $\max_j |P_{X,n_2}(B_j) - P_X(B_j)| \le \minprob/2 \Rightarrow \min_j P_{X,n_2}(B_j) \ge \minprob/2$ with probability $1-2\exp\{ -(\kappa_4/4) ( \minprob/J_\Fspace )^2 n_2 \}$ under the second inequality in~\eqref{eq:esup}. Hence adjust the definitions of $\kappa_3,\kappa_4$.

Thus it remains to show~\eqref{eq:esup}. For the first inequality, let $\{\zeta_i\}_i$ be independent Rademacher random
variables that are independent of $\{ X_i,Z_i \}_{i\in\samp_2}$. It
follows from the symmetrization Lemma 2.3.1 of~\citet{vaart} that $\left\| \sup_{\beta\in\Fspace^{\Psi_{n_2}}} |V_\beta| \right\|_{\Phi}$ is bounded by
$$\begin{aligned}
& 2\left\| n_2^{-1} \sup_{\beta\in\Fspace^{\Psi_{n_2}}} | \sum_{i\in\samp_2}\zeta_{i}\{l_{i}(F)-l_{i}(0)\} | \right\|_{\Phi} \\
\leq &
\frac{2}{n_2} \left\| \sup_{\beta\in\Fspace^{\Psi_{n_2}}} | \sum_{i\in\samp_2} \zeta_{i}
|Y_i'-\muhat(X_i)|^{\gamma} \left\{ e^{\beta(X_i)}-e^0\right\} | \right\|_{\Phi} + \frac{2}{n_2} \left\| \sup_{\beta\in\Fspace^{\Psi_{n_2}}} | \sum_{i\in\samp_2}\zeta_{i}\beta(X_{i}) | \right\|_{\Phi} \\
\leq &
\frac{2}{n_2}\Big\{ 2(3\gamma\bhi\log n_2)^\gamma e^{\Psi_{n_2}} + 1 \Big\}
\left\| \sup_{\beta\in\Fspace^{\Psi_{n_2}}} | \sum_{i\in\samp_2} \zeta_{i} \beta(X_i) | \right\|_{\Phi}
\leq
\tau_2 \frac{e^{\Psi_{n_2}}(\log n_2)^\gamma}{n_2} \left\| \sup_{\beta\in\Fspace^{\Psi_{n_2}}} \sum_{i\in\samp_2} \zeta_{i} \beta(X_i) \right\|_{\Phi}, \\
\end{aligned}$$
where the first term inside the curly brackets in the second inequality comes Theorem 4.12 of~\citet{ledoux}: The map $\beta\mapsto \{(3\gamma\bhi\log n_2)^\gamma e^{\Psi_{n_2}}\}^{-1} |Y_i'-\muhat(X_i)|^{\gamma}(e^{\beta}-e^0)$ is a contraction for $|\beta|<\Psi_{n_2}$ because $|e^{x}-e^{y}|\leq e^{\max(x,y)}|x-y|$ and because of~\eqref{eq:clip1}.
We drop the absolute value inside the supremum because $\Fspace^{\Psi_{n_2}}$ is closed under negation. Now hold $\{ X_i,Z_i \}_{i\in\samp_2}$ fixed so that only $\{\zeta_i\}_i$ is stochastic. By Corollary 2.2.7 and Theorem 2.2.4 of \citet{vaart},
$$
\begin{aligned}
\left\| \sup_{\beta\in\Fspace^{\Psi_{n_2}}} |V_\beta| \right\|_{\Phi}
&\le
\tau_2' \frac{e^{\Psi_{n_2}}(\log n_2)^\gamma}{n_2^{1/2}}
\int_0^{\Psi_{n_2}} \sqrt{ \log\mathcal{N}(u,\Fspace^{\Psi_{n_2}}, P_{n_2}) } du \le \tau_2' \frac{\Psi_{n_2} e^{\Psi_{n_2}}(\log n_2)^\gamma}{n_2^{1/2}} J_\Fspace \\
&= \tau_2' \frac{(\log n_2)^\gamma}{n_2^{1/10}}
\cdot
\frac{\Psi_{n_2} e^{\Psi_{n_2}}}{n_2^{1/5}}
J_\Fspace/n_2^{1/5} = \tau_2'' J_\Fspace/n_2^{1/5},
\end{aligned}
$$
where the second inequality follows from~\eqref{eq:entintegral}, and the last equality from $\Psi_{n_2}e^{\Psi_{n_2}} = n_2^{1/5}$ in Algorithm~\ref{alg:genericboost} and from the fact that $(\log n_2)^\gamma/n^{1/10} \le (10\gamma/e)^\gamma$ for $n_2 \ge 2$. The bound does not depend on $\{ X_i,Z_i \}_{i\in\samp_2}$, so the first inequality in~\eqref{eq:esup} holds. The second inequality in~\eqref{eq:esup} follows the same approach.
\hfill\Halmos
\endproof

\begin{lemma}\label{lem:lln}
For $\beta\in\{0,\betastar\}$, there exist positive constants $\kappa_5,\kappa_6$ depending on $\gamma$, $\blo$, and $\bhi$ for which
$$
\Prob\left( | R_n'(\muhat,\beta) - R(\muhat,\beta) | > \eta | \mathcal{W} \right)
\le
\kappa_5 \exp\left( -\kappa_6 \eta^2 n_2^{1/2} \right).
$$
\end{lemma}
\proof{Proof.}
Conditioning on $\mathcal{W}$, by~\eqref{eq:clip1} we have
$$
-\|\beta\|_\infty \le l(\beta) = |Y'-\muhat(X)|^{\gamma}e^{\beta(X)}-\beta(X) \le e^{\|\beta\|_\infty} (3\gamma\bhi\log n_2)^\gamma + \|\beta\|_\infty,
$$
from which it follows that the image of $l(\beta)$ is bounded by an interval no wider than $3e^{\|\beta\|_\infty} (3\gamma\bhi\log n_2)^\gamma$. Noting that $(\log n_2)^\gamma/n^{1/4} \le (4\gamma/e)^\gamma$ for $n_2\ge2$, we can further bound the interval width by $\tau_4 n_2^{1/4}$. Using Lemma~\ref{lem:cond_conc} and Hoeffding's inequality shows that for $\eta > \tau_3/n_2^{7/4}$,
$$
\begin{aligned}
\Prob\left( | R_n'(\muhat,\beta) - R(\muhat,\beta) | > \eta | \mathcal{W} \right)
&\le
\Prob\left( \left. \left|R_n'(\muhat,\beta) -\E(R_n'(\muhat,\beta)|\mathcal{W} )\right| > \eta-\tau_3/n_2^{7/4} \right| \mathcal{W} \right) \\
&\le
2\exp\left\{ -2n_2\left( \frac{ \eta-\tau_3/n_2^{7/4} }{ \tau_4 n_2^{1/4} } \right)^2 \right\}
\le
2\exp\left( \frac{(4\tau_3/\tau_4^2)\eta}{n_2^{5/4}}
- 2\eta^2/\tau_4^2 n_2^{1/2} \right) \\
&\le 
2\exp\left\{ \frac{(4\tau_3/\tau_4^2)}{3^{5/4}} - \tau_4'\eta^2 n_2^{1/2} \right\}
\le
\tau_3'''
\exp\left( -\tau_4'\eta^2 n_2^{1/2} \right).
\end{aligned}
$$
\hfill\Halmos
\endproof

\begin{lemma}\label{lem:gnd_tailbound}
For $W_1,\cdots,W_{n_2}$ independent draws from $GND(0,1)$,
$$
\Prob\left\{ \max_{i=1,\cdots,n_2}|W_i|>2\gamma\log n_2 \right\} < \gamma n_2^{-1},
\quad
\E\{ |W|^\gamma I_{(|W|>2\gamma\log n_2)} \} \le \frac{6\log n_2}{n_2^2}.
$$
\end{lemma}
\proof{Proof.}
GND has the fattest tails (exponential) when $\gamma=1$. For $\kappa>1$,
$$
\Prob(|W_i|>\kappa) = \frac{2}{2\gamma^{1/\gamma}\Gamma(1+\gamma^{-1})} \int_{\kappa}^\infty e^{-w^\gamma/\gamma} dw \le \int_{\kappa}^\infty e^{-w/\gamma} dw = \gamma e^{-\kappa/\gamma},
$$
where the inequality holds for $\gamma\ge 1$. Hence $\Prob( |W_i|>2\gamma\log n_2 ) \le \gamma n_2^{-2}$ and the first claim follows from the union bound.
For the second claim, the second inequality below relies on the fact that $2\gamma\log n_2 \ge 1$:
$$
\begin{aligned}
\E\{ |W|^\gamma I_{(|W|>2\gamma\log n_2)} \} &= \frac{2}{2\gamma^{1/\gamma}\Gamma(1+\gamma^{-1})} \int_{2\gamma\log n_2}^\infty w^\gamma e^{-w^\gamma/\gamma} dw \le \gamma^{1/\gamma} \int_{(2\gamma\log n_2)^\gamma/\gamma}^\infty t^{1/\gamma} e^{-t} dt \\
&\le 2 \int_{(2\gamma\log n_2)/\gamma}^\infty t^{1/\gamma} e^{-t} dt \le 2 \int_{2\log n_2}^\infty t e^{-t} dt = 2\frac{1+2\log n_2}{n_2^2} \le 6\frac{\log n_2}{n_2^2}.
\end{aligned}
$$
\hfill\Halmos
\endproof

\begin{lemma}\label{lem:cond_conc}
For $\beta \in \{0,\betastar\}$, there exist positive constants $\tau_1,\tau_3$ depending on $\gamma$, $\blo$, and $\bhi$ for which
$$
\Prob( U_{\Fspace^{\Psi_{n_2}}} | \mathcal{W} )
\le
\Prob\left(\left. \sup_{\beta\in\Fspace^{\Psi_{n_2}}} |V_\beta| >\eta-\tau_1/n_2^{3/2} \right| \mathcal{W} \right),
\qquad
\Prob( U_{\beta} | \mathcal{W} )
\le
\Prob\left(\left. |V_\beta| >\eta-\tau_3/n_2^{7/4} \right| \mathcal{W} \right).
$$
\end{lemma}
\proof{Proof.}
For $\Prob\left(\left. U_{\Fspace^{\Psi_{n_2}}} \right| \mathcal{W} \right)$ we will show that $\sup_{\beta\in\Fspace^{\Psi_{n_2}}} \Big| R(\muhat,\beta)-\E(R_n'(\muhat,\beta)|\mathcal{W}) \Big| \le \tau_1/n_2^{3/2}$, in which case
$$
\begin{aligned}
|\{R_n'(\muhat,\beta)-R_n'(\muhat,0)\} - \{R(\muhat,\beta)-R(\muhat,0)\} | &\le |V_\beta| + 2\sup_{\beta\in\Fspace^{\Psi_{n_2}}} \Big| R(\muhat,\beta)-\E(R_n'(\muhat,\beta)|\mathcal{W}) \Big| \\
&\le |V_\beta| + 2\tau_1 /n_2^{3/2},
\end{aligned}
$$
so we can redefine $\tau_1$ for the first claim. Turning to the bound for $\sup_{\beta\in\Fspace^{\Psi_{n_2}}} \Big| R(\muhat,\beta)-\E(R_n'(\muhat,\beta)|\mathcal{W}) \Big|$, note from~\eqref{eq:cond_exp} that
$$
\begin{aligned}
\E \{R_n'(\muhat, \beta)|\mathcal{W}\} &= n_2^{-1}\sum_{i\in\samp_2} \E_{X_i} \left\{ e^{\beta(X_i)} \int_{|W_i|\le 2\gamma\log n_2} |Y_i'-\muhat(X_i)|^{\gamma} \frac{dP_{W_i}}{1-Q_{n_2}} - \beta(X_i) \right\} \\
&= (1-Q_{n_2})^{-1} \E_X \left\{ e^{\beta(X)}\int_{|W|\le 2\gamma\log n_2} |Y'-\muhat(X)|^{\gamma} dP_{W} - (1-Q_{n_2})\beta(X) \right\} \\
&= (1-Q_{n_2})^{-1} R(\muhat,\beta) \\
&\qquad +(1-Q_{n_2})^{-1}\E_X \left\{ -e^{\beta(X)}\int_{|W|> 2\gamma\log n_2} |Y'-\muhat(X)|^{\gamma} dP_{W} + Q_{n_2}\beta(X) \right\},
\end{aligned}
$$
where $Q_{n_2} = \Prob (|W|> 2\gamma\log n_2) \le \gamma n_2^{-2}$ from the analysis in Lemma~\ref{lem:gnd_tailbound}. Rearranging terms yield
$$
\begin{aligned}
R(\muhat,\beta) - \E\{R_n'(\muhat,\beta)|\mathcal{W}\}  &= -Q_{n_2} \E\{R_n'(\muhat,\beta)|\mathcal{W}\} -  Q_{n_2}\E_X\beta(X) \\
&\qquad 
+\E_X \left\{ e^{\beta(X)}\int_{|W|> 2\gamma\log n_2} |Y_i'-\muhat(X)|^{\gamma} dP_{W}  \right\}.
\end{aligned}
$$
On $\mathcal{W}$, $|R_n'(\muhat,\beta)|\le e^{\Psi_{n_2}} (3\gamma\bhi\log n_2)^\gamma + \Psi_{n_2}$ by~\eqref{eq:clip1}. 
Furthermore, applying Jensen's inequality to $|\cdot|^\gamma$ shows that $|Y'-\muhat(x)|^\gamma < 2^\gamma|Y'-\mu(x)|^\gamma + 2^\gamma\e_{n_1}^\gamma \le 2^\gamma|Y'-\mu(x)|^\gamma + (2\gamma\bhi)^\gamma$, so
$$
\begin{aligned}
\Big| R(\muhat,\beta) - \E\{R_n'(\muhat,\beta)|\mathcal{W}\} \Big|
&<
(3\gamma\bhi)^\gamma e^{\Psi_{n_2}}(\log n_2)^\gamma Q_{n_2} + 2\Psi_{n_2} Q_{n_2} \\
&
\quad + 2^{\gamma}e^{\Psi_{n_2}} \E_X \int_{|W|> 2\gamma\log n_2} |Y_i'-\mu(X)|^{\gamma} dP_{W} + (2\gamma\bhi)^\gamma e^{\Psi_{n_2}} Q_{n_2} \\
&\le
3(3\gamma\bhi)^\gamma e^{\Psi_{n_2}}(\log n_2)^\gamma Q_{n_2} + (2\bhi)^\gamma e^{\Psi_{n_2}} \int_{|w|> 2\gamma\log n_2} |w|^\gamma dP_{W} \\
&\le
3\gamma(3\gamma\bhi)^\gamma \frac{e^{\Psi_{n_2}}(\log n_2)^\gamma}{n_2^2} + 6(2\bhi)^\gamma \frac{e^{\Psi_{n_2}}\log n_2}{n_2^2} \\
&\le
9\gamma(3\gamma\bhi)^\gamma \frac{e^{\Psi_{n_2}}(\log n_2)^\gamma}{n_2^2}
=
9\gamma(3\gamma\bhi)^\gamma \frac{e^{\Psi_{n_2}}}{n_2^{1/4}} \cdot \frac{(\log n_2)^\gamma}{n_2^{1/4}}/n_2^{3/2}
\le
\tau_1/n_2^{3/2},
\end{aligned}
$$
where the third inequality follows from Lemma~\ref{lem:gnd_tailbound}. The last inequality comes from the definition of $\Psi_{n_2}$ in Algorithm~\ref{alg:genericboost} and the fact that $(\log n_2)^\gamma/n_2^{1/4} \le (4\gamma/e)^\gamma$ for $n_2\ge2$. This establishes the first claim.

For the second claim, a similar argument shows that for $\beta\in\{0,\betastar\}$,
$$
\begin{aligned}
\Big| R(\muhat,\beta) - \E\{R_n'(\muhat,\beta)|\mathcal{W}\} \Big|
&\le 
|V_\beta| + 18\gamma(3\gamma\bhi)^\gamma e^{\|\beta\|_\infty} \frac{(\log n_2)^\gamma}{n_2^2} \\
&= 
|V_\beta| + \tau_3 e^{\|\beta\|_\infty} \frac{(\log n_2)^\gamma}{n_2^{1/4}}/n_2^{7/4} 
\le 
|V_\beta| + \tau_3' e^{\|\beta\|_\infty}/n_2^{7/4}.
\end{aligned}
$$
Clearly, $e^{\|0\|_\infty}=1$. Moreover, from Lemma~\ref{lem:betastar} we have $\| \betastar \|_\infty \le \gamma \max\{ \log(2/\blo), \log(2\bhi) \}$ hence $e^{\| \betastar \|_\infty} \le \max\left\{ (2/\blo)^{\gamma}, (2\bhi)^{\gamma} \right\}$. Thus we can absorb $\max( e^{\|0\|_\infty}, e^{\|\betastar\|_\infty} )$ into the definition of $\tau_3'$ and relabel as $\tau_3$ to establish the second claim.
\hfill\Halmos
\endproof

\subsubsection{Minimization of empirical risk.}\label{subsec:minrisk}

\begin{lemma}\label{lem:bhat_bound}
Let $n_2 \ge 3$ and $ \| \mu-\muhat \|_\infty = \e_{n_1} \le \blo/2$ in~\eqref{eq:muhat_property}. Then there exist constants $\kappa_1,\kappa_2,\kappa_3$ depending on $\gamma$, $\blo$, and $\bhi$ such that with probability at least $1 - \kappa_1 \exp\left\{ - \kappa_2 (\minprob /J_\Fspace)^2 n_2^{2/5} \right\} - \gamma n_2^{-1}$, the boosting iterates $\beta_0=0, \beta_1, \cdots, \beta_{\hat m} = \betahat$ as well as $\betastar$ are all uniformly bounded, i.e. $\| \beta_m \|_\infty, \| \betastar \|_\infty \le \minprob^{-1} \kappa_3$.
\end{lemma}
\proof{Proof.}
Under the stated hypotheses, setting $\eta=1$ in Proposition~\ref{prop:main_conc} implies that $R(\muhat,\beta_m) \le R_n(\muhat,\beta_m) + 2\eta \le R_n(\muhat,0) + 2\eta \le R(\muhat,0) + 3\eta \le (2\bhi)^\gamma + 3$ with the stated probability. The last inequality is because $R(\muhat,0) = \int \bstar(x)^\gamma dP_x \le (2\bhi)^\gamma$ by Lemma~\ref{lem:betastar}. 

For a fixed scalar $s>0$, the function $s e^\beta - \beta$ is bounded below by $-\beta$ and by $\beta+2\{1-\log(2/s)\}$, and hence by $|\beta| + 2\min\{0,1-\log(2/s)\}$. It follows from~\eqref{eq:pop_risk} and the first result in Lemma~\ref{lem:betastar} that
\begin{equation}\label{eq:poprisk_LB}
\begin{aligned}
R(\muhat,\beta) &= \sum_j \left\{
(\bstar|_{B_j})^\gamma e^{\beta|_{B_j}} - \beta|_{B_j} \right\}  P_X(B_j)
\ge \sum_j 
\left[
\left|\beta|_{B_j} \right| + 2\min\{0,1-\log(2^{\gamma+1}/\blo^\gamma)\}
\right] P_X(B_j) \\
&\ge \minprob \| \beta \|_\infty + 2\min\{0,1+\gamma\log(\blo/4)\},
\end{aligned}
\end{equation}
where we used the bound $\bstar(x) \ge \blo/2$ from Lemma~\ref{lem:betastar} in the first inequality. 
As shown earlier, $R(\muhat,\beta_m) \le (2\bhi)^\gamma + 3$ with high probability, so $\kappa_3 = (2\bhi)^\gamma + 3 - 2\min\{0,1+\gamma\log(\blo/4)\}$. From Lemma~\ref{lem:betastar} we also have $\| \betastar \|_\infty \le \gamma \max\{ \log(2/\blo), \log(2\bhi) \}$, so let $\kappa_3$ be the maximum of these two.
\hfill\Halmos
\endproof

For the next result, recall from Algorithm~\ref{alg:genericboost} in Appendix~\ref{sec:alg2} that the empirical inner product and norm are $\langle f,f' \rangle = n_2^{-1} \sum_{i\in\samp_2} f(x_i)f'(x_i)$ and $\| f \| = \langle f,f \rangle^{1/2}$.
Recall also that $\varepsilon$ is fixed between $(0,1]$, $\Psi_{n_2}\nearrow\infty$ and $\nu_{n_2}^2(\log n_2)^\gamma \searrow 0$ as $n_2\rightarrow\infty$,
so the result implies that $R_n(\muhat,\betahat)-R_n(\muhat,\betastar) = o_p(1)$.
\begin{proposition}\label{prop:minrisk}
Let $n_2 \ge 3$ and $\| \mu-\muhat \|_\infty = \e_{n_1} \le \blo/2$ in~\eqref{eq:muhat_property}. Then there exist constants $\kappa_1,\kappa_2,\kappa_3,\kappa_4$ depending on $\gamma$, $\blo$, and $\bhi$ such that with probability at least $1 - \kappa_1 \exp\left\{ - \kappa_2 (\minprob /J_\Fspace)^2 n_2^{2/5} \right\} - \gamma n_2^{-1}$,
$$
R_n(\muhat, \betahat) - R_n(\muhat, \betastar)
<
\kappa_3 \exp\left( -\frac{\varepsilon\minprob^2}{4\kappa_3}\Psi_{n_2} \right)
+
e^{\kappa_4/\minprob} \nu_{n_2}^2(\log n_2)^{\gamma}.
$$
\end{proposition}
\proof{Proof.}
We adapt the proof of Lemma 2 in \citet{lee2021boosted} for the empirical risk~\eqref{eq:emp_risk}. By~\eqref{eq:clip1} and Lemma~\ref{lem:gnd_tailbound}, $\max_{i\in\samp_2} |y_i-\muhat(X_i)|^\gamma \le (3\gamma\bhi\log n_2)^\gamma$ with the probability stated in Proposition~\ref{prop:minrisk}. Recalling from Lemma~\ref{lem:bhat_bound} that the boosting iterates are uniformly bounded by $\kappa_3/\minprob$ with the same probability, we obtain the following Taylor bound for~\eqref{eq:emp_risk}:
\begin{equation}\label{eq:emp_taylor}
\begin{aligned}
R_n(\muhat,\beta_{m+1}) &\le R_n(\muhat,\beta_m) - \frac{\varrho_m\nu_{n_2}}{m+1}\langle g_{\beta_m}, g^\varepsilon_{\beta_m} \rangle + \frac{\varrho_m^2\nu_{n_2}^2 (\log n_2)^{\gamma}}{2(m+1)^2} (3\gamma\bhi)^\gamma e^{\kappa_3/\minprob} \| g^\e_{\beta_m} \|^2 \\
&\le R_n(\muhat,\beta_m) - \frac{\varrho_m\nu_{n_2}}{m+1} \varepsilon \| g_{\beta_m} \| + \frac{\nu_{n_2}^2 (\log n_2)^{\gamma}}{2(m+1)^2} e^{\kappa_4/\minprob},
\end{aligned}
\end{equation}
where the last inequality uses~\eqref{eq:eps_grad} and the fact that $\|g^\varepsilon_{\beta_m}\|=1$. Enlarge $\kappa_3$ to $\kappa_4$ to absorb $(3\gamma\bhi)^\gamma$.

We may assume that $R_n(\muhat,\beta_{\hat m})-R_n(\muhat, \betastar) > 0$, i.e. $\beta_{\hat m}$ is not the minimizer of the empirical risk, for otherwise the proposition automatically holds. Thus Algorithm~\ref{alg:genericboost} terminates with either $\hat m = \infty$ or $\| \beta_{\hat m} - \frac{\varrho_{\hat m}\nu_{n_2}}{\hat m+1} g^\varepsilon_{\beta_{\hat m}} \|_\infty \ge \Psi_{n_2}$, in which case 
\begin{equation}\label{eq:sum_steps}
\begin{aligned}
\Psi_{n_2}
\le 
\sum_{m=0}^{\hat m} \frac{\varrho_m\nu_{n_2}}{m+1}  \| g^\varepsilon_{\beta_m} \|_\infty
\le
\frac{2}{\minprob} \left( \sum_{m=0}^{\hat m-1} \frac{\varrho_m\nu_{n_2}}{m+1} + 1 \right)
\end{aligned}    
\end{equation}
with the stated probability in the proposition. The last inequality is because for $f = \sum_j f_j I_{B_j}(x) \in\Fspace$, $\|f\|^2 \ge \| f \|_\infty \cdot \min_j P_{X,n_2}(B_j)$, and $\min_j P_{X,n_2}(B_j) \ge \minprob/2$ via Proposition~\ref{prop:main_conc}. Recall also that $\| g^\varepsilon_{\beta_m} \|=1$.

By convexity, $R_n(\muhat,\betastar)-R_n(\muhat,\beta_m) \ge \langle g_{\beta_m}, \betastar - \beta_m \rangle$. Also, $\langle g_{\beta_m}, \beta_m - \betastar \rangle \le \| g_{\beta_m} \| \cdot \| \beta_m-\betastar \| \le 2\minprob^{-1}\kappa_3 \| g_{\beta_m} \|$ by Lemma~\ref{lem:bhat_bound}. Putting these into~\eqref{eq:emp_taylor}, and subtracting $R_n(\muhat,\betastar)$ from both sides of the inequality shows that for $\delta_m = R_n(\muhat,\beta_m)-R_n(\muhat,\betastar)$,
$$
\begin{aligned}
\delta_{m+1} &\le \left( 1 - \frac{\varepsilon\minprob}{2\kappa_3} \frac{\varrho_m\nu_{n_2}}{m+1} \right) \delta_m + \frac{\nu_{n_2}^2(\log n_2)^{\gamma}}{2(m+1)^2} e^{\kappa_4/\minprob}.
\end{aligned}
$$
Replacing $\kappa_3$ with $\max\{1,\kappa_3\}$ if necessary to ensure that the term inside the parenthesis is between 0 and 1, solving the recurrence yields
$$
\begin{aligned}
\delta_{\hat m} &\le \delta_0 \prod_{m=0}^{\hat m-1}
\left( 1 - \frac{\varepsilon\minprob}{2\kappa_3} \frac{\varrho_m\nu_{n_2}}{m+1} \right)
+
e^{\kappa_4/\minprob} \nu_{n_2}^2(\log n_2)^{\gamma} \sum_{m=0}^\infty \frac{1}{2(m+1)^2} \\
&< e \max(0,\delta_0) \exp\left( -\frac{\varepsilon\minprob^2}{4\kappa_3}\Psi_{n_2} \right)
+
e^{\kappa_4/\minprob} \nu_{n_2}^2(\log n_2)^{\gamma},
\end{aligned}
$$
where we used~\eqref{eq:sum_steps} and the fact that $1-y \le e^{-y}$ for $|y|<1$. Finally, to complete the claim, we need to show that $\delta_0\le \kappa_3$, and then enlarge $\kappa_3$ to absorb the extra factor $e$ above. Proposition~\ref{prop:main_conc} shows that with the probability stated in this proposition, $\delta_0 \le R(\muhat,0) - R(\muhat,\betastar) + 2 \le (2\bhi)^\gamma + 2 - R(\muhat,\betastar) \le (2\bhi)^\gamma + 2 - 2\min\{0,1+\gamma\log(\blo/4)\} < \kappa_3$, where the second inequality comes from $R(\muhat,0) = \int \bstar(x)^\gamma dP_x \le (2\bhi)^\gamma$ by Lemma~\ref{lem:betastar}, and the third one from~\eqref{eq:poprisk_LB} in Lemma~\ref{lem:bhat_bound}.
\hfill\Halmos
\endproof

\section{Forecasting ED Wait and Service Times with \bgnd}\label{sec:results}

We now turn to the ED forecasting problem that motivated the development of \bgnd. We show that combining distributional knowledge from the operations literature with the flexibility of ML leads to distributional forecasts that are superior to both the distribution-agnostic machine learning QRF and operations-informed parametric models from classical statistics.

Our data come from a large academic ED in the U.S. and contains over 190,000 patient visit records between 2016 and 2019 inclusive. For each visit, we have information on the patient's gender, age, race, category of the main complaint, and timestamps for activities including arrival, departure, and service start times. The day of week is transformed into $\{ \sin(\frac{2\pi t}{24\times7}), \cos(\frac{2\pi t}{24\times7}) \}$ to capture the weekly cycle, and the time of arrival into $\{ \sin(\frac{2\pi t}{24}), \cos(\frac{2\pi t}{24}) \}$ to capture the daily cycle. We use these as predictors because it is known that customer arrival patterns can be well modelled by a nonhomogeneous Poisson process with a sinusoidal arrival rate function \citep{chen2019super, chen2024can}.

The typical patient in our dataset is female (60\%), African American (56\%), 50 years old (median), and has a primary complaint that falls into the general medicine category (19\%). The median wait time is 52 minutes and the median service time is 4 hours and 24 minutes. Wait times are particularly long between 3 and 6 pm, while service times are longest for patients who begin receiving care between 6 am and 9 am. Detailed summary statistics are presented in the electronic companion.

We generate distributional forecasts for visits in the 2017 test period by training models on 2016 data, and we repeat the process for the 2018 and 2019 test periods using the preceding year as the training period.
The quality of the forecasts is measured with CRPS and patient outcomes, and QRF is used as our benchmark.

First, we compare the accuracy of alternative approaches in forecasting wait time distributions. Table \ref{tab:crps} in the Introduction presents the improvements in CRPS relative to QRF. As discussed in Section~\ref{subsec:ops_value}, the classic exponential model, which is informed by operations knowledge, provides an average improvement of 2.5\% for forecasting ED wait times. \bgnds achieves an even bigger improvement of 6.1\%. For forecasting service time distributions, the classic log-normal model achieves an improvement of 7.0\%, while \bgnds achieves 8.8\%.


Further analysis suggests that these improvements in forecasting also translate into better patient outcomes (see the electronic companion for details). Table \ref{tab:applications}  summarizes these results. First, wait time announcements based on \bgnds forecasts reduce patient dissatisfaction associated with waiting by 9.4\% when patients are loss averse, as demonstrated in \citet{ansariUnderPromising2022a}. Second, our findings suggest that \bgnds can potentially save 3,300 lives per year amongst ED patients presenting with myocardial infarction, or 4.1\% more than QRF. This potential reduction corresponds to lives that could be saved by correctly identifying cardiac arrest patients most likely to miss treatment within the \textit{golden hour} \citep{boersmaEarly1996}.

\begin{table}
\SingleSpacedXI
\small

\caption{\label{tab:applications}Improvements in patient satisfaction and mortality}
\begin{threeparttable}
\centering
\makebox[\linewidth]
{\begin{tabular}[]{ccccc}
\toprule
  & \bgnds & QRF & Classic Exponential & Improvement\\
\midrule
Quantile loss & 0.50 & 0.55 & 0.53 & 9.4\%\\
Reduction in mortality & 3,300 & 3,100 & 3,100 & 4.1\%\\
\bottomrule
\end{tabular}}
\begin{tablenotes}
\small
\item \textit{Notes: } Quantile loss, used to model patient dissatisfaction, is calculated at the 70th percentile \citep{ansariUnderPromising2022a}. The mortality reduction shows the number of mortalities that can be prevented by identifying patients who are most likely to miss treatment within the golden hour. The improvement column is \bgnd's percentage improvement relative to QRF in terms of quantile loss and prevented mortalities. All figures are rounded to two significant digits. See the electronic companion for details.
\end{tablenotes}
\end{threeparttable}
\end{table}

\section{Discussion}

We address critical gaps in the application of ML techniques to operational settings by introducing the boosted Generalized Normal Distribution (\bgnd), which combines the flexibility of ML with distributional knowledge from the operations literature. The \bgnds not only renders more efficient and accurate distributional forecasts, which is essential for solving fundamental operations problems, but the formal guarantees we provide is a first for the recent literature on parametric machine learning with gradient boosting. It is also worth noting that our guarantees extend to all special cases of the \bgnds studied in the ML literature. Similar efforts to constrain the search space of ML algorithms to reflect the underlying mechanism have been shown in other fields such as computational physics \citep{raissi2019physics}.


Our application to forecasting patient wait and service times at a large academic ED in the U.S. demonstrates the practical utility of \bgnd. We highlight the role of operations-specific knowledge by comparing QRF against \bgnd, and against operations-informed parametric models from classical statistics. While the latter is already enough to outperform QRF, \bgnds does even better by combining ML with operations knowledge. Our analysis suggests that these improved forecasts lead to higher patient satisfaction ($+9.4\%$) and reduced cardiac arrest mortality rates ($-4.1\%$). Beyond the ED, \bgnds can be applied to tackle other challenges in healthcare operations such as surgery scheduling, hospital staffing, or ambulance siting \citep{westgate2013travel}, all problems where distributional forecasts are essential. \bgnds can also be applied to forecast healthcare expenditures for households, for which the distribution of expenditures appear to follow a gamma distribution \citep{lowsky2018health}, which can also be well approximated by GND.


The rise of ML approaches in solving operations problems have overshadowed some key findings in the literature. Our paper demonstrates the value of these findings, showing that when combined with classical forecasting models, they can outperform ML techniques. Furthermore, we introduce \bgnd, an approach that integrates operations knowledge with machine learning, resulting in even greater performance. \bgnds is statistically consistent, no more difficult to use than off-the-shelf ML algorithms, and also outperforms them. There is every reason to deploy it in practice. 

\counterwithin{equation}{section}
\begin{APPENDICES}
\section{Boosted estimation of log-scale parameter $\beta(x)$}\label{sec:alg2}

Algorithm~\ref{alg:genericboost} below follows the structure of Algorithm 1 in \citet{lee2021boosted}, and estimates $\betahat(x)$ with gradient-boosted trees. This is done using regularized gradient descent to minimize the empirical risk $R_n(\muhat,\beta,\samp_2)$, resulting in the iterates $\beta_0=0, \beta_1, \cdots, \beta_{\hat m} = \betahat$. The main regularizations used are: i) Stopping the iterations early so that the uniform norm of $\beta_m$ stay below the limit $\Psi_{n_2}$ defined below; and ii) Restricting the stepsize of the update from $\beta_m$ to $\beta_{m+1}$ to be less than $\nu_{n_2}/(m+1)$, where $\nu_{n_2}$ satisfies~\eqref{eq:shrinkage}. Additionally, in boosting the iterates are not usually updated in the exact direction of the negative gradient $-g_{\beta_m}$ of $R_n(\muhat,\beta_m,\samp_2)$. Rather, $g_{\beta_m}$ is first approximated by a tree learner $g_{\beta_m}^{\varepsilon}$ of limited depth, and $-g_{\beta_m}^{\varepsilon}$ is then used as the direction of update. The degree of approximation depends on the depth of the tree used and is characterized in terms of the correlation $\varepsilon\in(0,1]$ between $g_{\beta_m}^{\varepsilon}$ and $g_{\beta_m}$ in~\eqref{eq:eps_grad}. In practice it is common to use $K$-fold cross-validation to select the number of iterations and tree depth. We follow this approach in our empirical analysis with $K=10$. Finally, for functions $f,f'\in\Fspace$, the empirical inner product and 2-norm appearing in the algorithm are $\langle f,f' \rangle = n_2^{-1} \sum_{i\in\samp_2} f(x_i)f'(x_i)$ and $\| f \| = \langle f,f \rangle^{1/2}$ respectively. 

\begin{algorithm}[htpb]
\centering

\caption{Boosted estimation of log-scale parameter $\beta(x)$\label{alg:genericboost}}

\begin{algorithmic}
\SingleSpacedXI
\STATE Initialize $\beta_0=0$, $m=0$; set $\varepsilon\in(0,1]$, $\Psi_{n_2} = W(n_2^{1/5}) \rightarrow \infty$ where $W(y)e^{W(y)} = y$, and choose a stepsize shrinkage $\nu_{n_2}>0$ to satisfy
\vspace{-8pt}
\begin{equation}\label{eq:shrinkage}
\nu_{n_2}<1,
\qquad
\nu_{n_2}^2(\log n_2)^\gamma \searrow 0.
\end{equation}
\vspace{-12pt}
\WHILE {gradient $g_{\beta_m} = 
\sum_j \left\{
\frac{ \sum_{i\in\samp_2} \left| y_i-\muhat|_{B_j} \right|^\gamma }{n_2} e^{\beta_m|_{B_j}}
- P_{X,n_2}(B_j) 
\right\} I_{B_j}(x) \ne 0$}
\STATE Compute an $\varepsilon$-gradient $g_{\beta_m}^{\varepsilon}\in\Fspace$ of unit length ($\| g_{\beta_m}^{\varepsilon} \| = 1$) satisfying
\begin{equation}\label{eq:eps_grad}
\left\langle
g_{\beta_m}/\|g_{\beta_m}\| ,g_{\beta_m}^{\varepsilon}
\right\rangle \geq \e
\end{equation}
\vspace{-15pt}
\STATE Compute $\beta\leftarrow \beta_m-\frac{\varrho_m\nu_{n}}{m+1}g_{\beta_m}^{\varepsilon}$
where $\varrho_m = \arg\min_{\varrho\in(0,1]} R_n(\muhat,\beta_m-\frac{\varrho\nu_{n}}{m+1}g_{\beta_m}^{\varepsilon},\samp_2)$
\IF {$\| \beta \|_{\infty} < \Psi_{n_2}$}
\STATE Update the log-scale estimator:
$\beta_{m+1}\leftarrow \beta$
\STATE Update $m\leftarrow m+1$
\ELSE
\STATE \textbf{break}
\ENDIF
\ENDWHILE
\STATE Set $\mhat\leftarrow m$. The log-scale estimator is
$\betahat = \beta_{\mhat}= -\sum_{m=0}^{\mhat-1}\frac{\varrho_m\nu_{n_2}}{m+1} g_{\beta_m}^\varepsilon$
\end{algorithmic}
\end{algorithm}

\end{APPENDICES}

\clearpage
\newpage
\bibliographystyle{informs2014} 
\bibliography{refs} 

\ECSwitch


\ECHead{Electronic Companion for ``Boosted Generalized Normal Distributions: Integrating Machine Learning with Operations Knowledge''}

\section{Example of non-convexity of the expected negative log-likelihood surface}\label{sec:non-convex}

The expected negative log-likelihood (i.e. population risk) is the limit of the empirical risk as the sample size approaches infinity. Analyzing the minimization of the population risk is simpler than analyzing empirical risk minimization because noise is absent from the former setting (as the entire data distribution is known). We provide an example below where even in the limit of infinite sample size, the surface of the population risk is non-convex. While non-convexity does not necessarily prevent analyzing the statistical consistency of risk minimization via gradient descent, it adds a layer of complexity that may explain why existing algorithms have not been analyzed.

For our example we consider the simple case where there are no covariates, and $Y \sim N(\mu_0,\sigma_0^2)$. Suppose the unknown true mean is in fact $\mu_0=0$ and the unknown true standard deviation is $\sigma_0=1$. For candidate values $(\mu,\sigma)$ of the mean and standard deviation, the expected negative log-likelihood is, modulo a constant,
$$
R(\mu,\sigma)
=
\E_{Y\sim N(\mu_0,\sigma_0^2)}
\left\{
\log\sigma + \frac{(Y-\mu)^2}{2\sigma^2}
\right\}
=
\log\sigma + \frac{1+\mu^2}{2\sigma^2}.
$$
To see that the surface of $R(\mu,\sigma)$ is non-convex, note that its Hessian at $(\mu,\sigma) = (2,1)$ is
$$
\left(
\begin{array}{cc}
1 & -4 \\
-4 & 14
\end{array}
\right),
$$
which has a negative eigenvalue of $(15-\sqrt{233})/2$. The surface remains non-convex even if we reparameterize the distribution in terms of log-standard deviation instead of $\sigma$.

\section{Descriptive Statistics}

\begin{figure}[h]
    \centering

    \begin{subfigure}[t]{1\textwidth}
    \centering
    \includegraphics[width=0.9\textwidth]{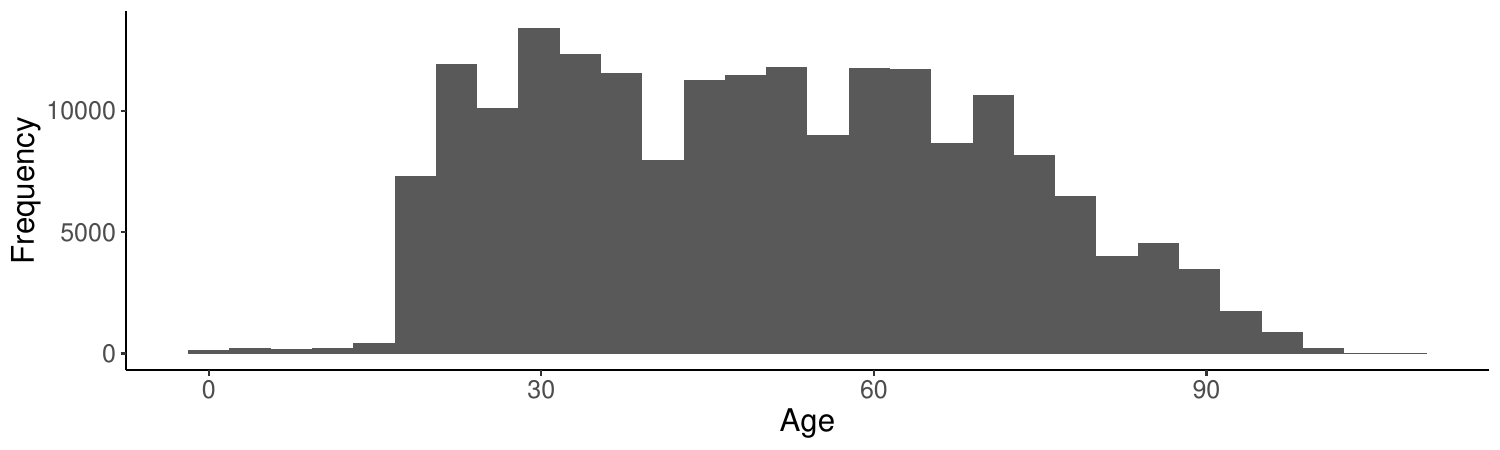} 
    \caption{Histogram of patient age} 
    \end{subfigure}
    
    \begin{subfigure}[t]{0.9\textwidth}
    \centering
    \includegraphics[width=\textwidth]{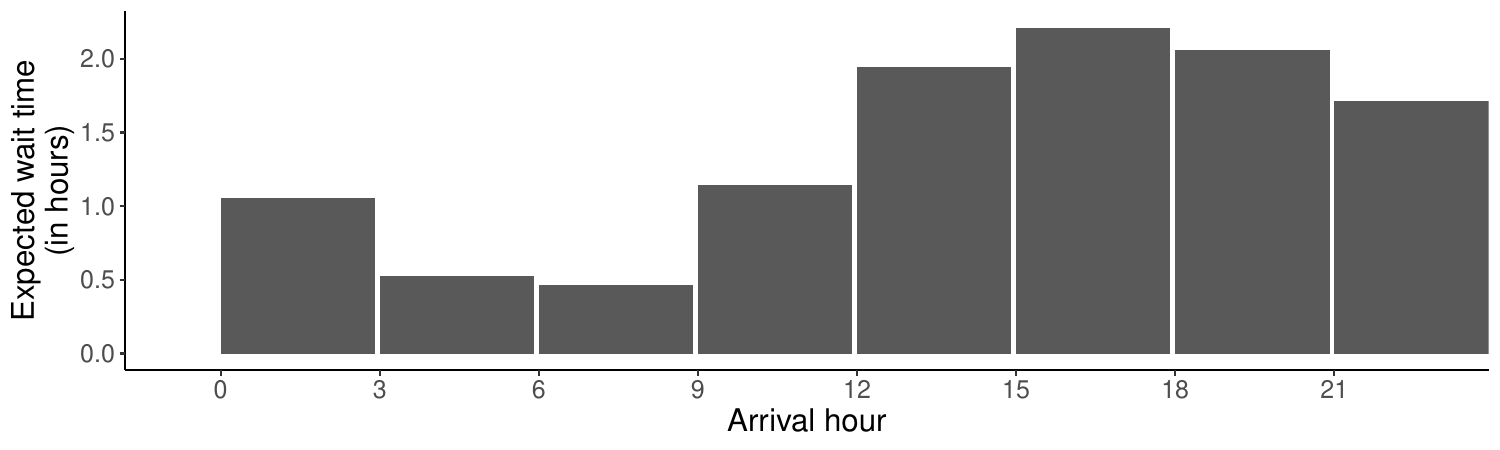} 
    \caption{Expected wait time throughout a day} 
    \end{subfigure}
    
    \begin{subfigure}[t]{0.9\textwidth}
    \centering
    \includegraphics[width=\textwidth]{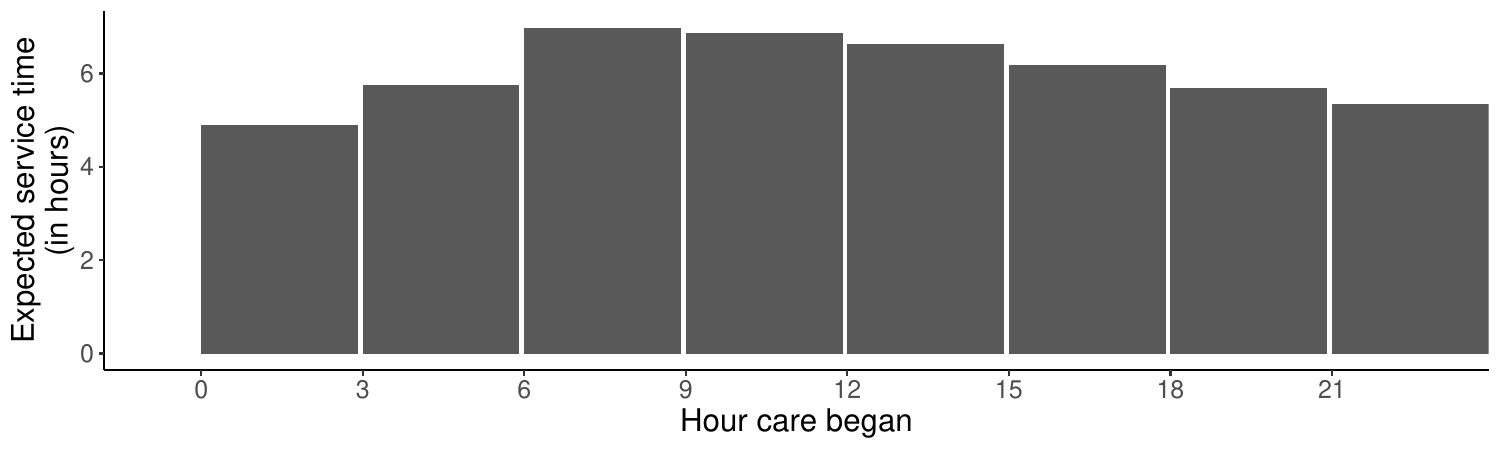} 
    \caption{Expected service time throughout a day} 
     \end{subfigure}

    \caption{Summary of the numeric variables}
    \label{fig:num_stats}
\end{figure}

We present summary statistics of our data in Table~\ref{tab:cat_stats} and Figure~\ref{fig:num_stats}. While Table~\ref{tab:cat_stats} shows the proportions of the categories observed in the data, Figure~\ref{fig:num_stats} presents the histogram of patient age and how wait and service times vary over time of day.

\begin{table}

\caption{\label{tab:cat_stats}Summary of the categorical variables}
\centering
\begin{tabular}[t]{rr}
\toprule
Category & Proportion\\
\midrule
\addlinespace[0.3em]
\multicolumn{2}{l}{\textbf{Gender}}\\
\hspace{1em}Female & 60\%\\
\hspace{1em}Male & 40\%\\
\addlinespace[0.3em]
\multicolumn{2}{l}{\textbf{Race}}\\
\hspace{1em}African American  or Black & 56\%\\
\hspace{1em}American Indian or Alaskan Native & 0.25\%\\
\hspace{1em}Asian & 3.0\%\\
\hspace{1em}Caucasian or White & 39\%\\
\hspace{1em}Hispanic & $<0.1$\%\\
\hspace{1em}Multiple & 0.35\%\\
\hspace{1em}Native Hawaiian or Other Pacific Islander & 0.11\%\\
\hspace{1em}Unknown, Unavailable or Unreported & 1.5\%\\
\addlinespace[0.3em]
\multicolumn{2}{l}{\textbf{Main complaint category}}\\
\hspace{1em}Cardio & 17\%\\
\hspace{1em}Gastro & 13\%\\
\hspace{1em}General & 19\%\\
\hspace{1em}Neurology & 5.6\%\\
\hspace{1em}Non-Traumatic & 6.3\%\\
\hspace{1em}Other & 32\%\\
\hspace{1em}Trauma & 6.9\%\\
\bottomrule
\end{tabular}
\end{table}

\begin{figure}[h!]
    \centering
    \includegraphics[width=1\textwidth]{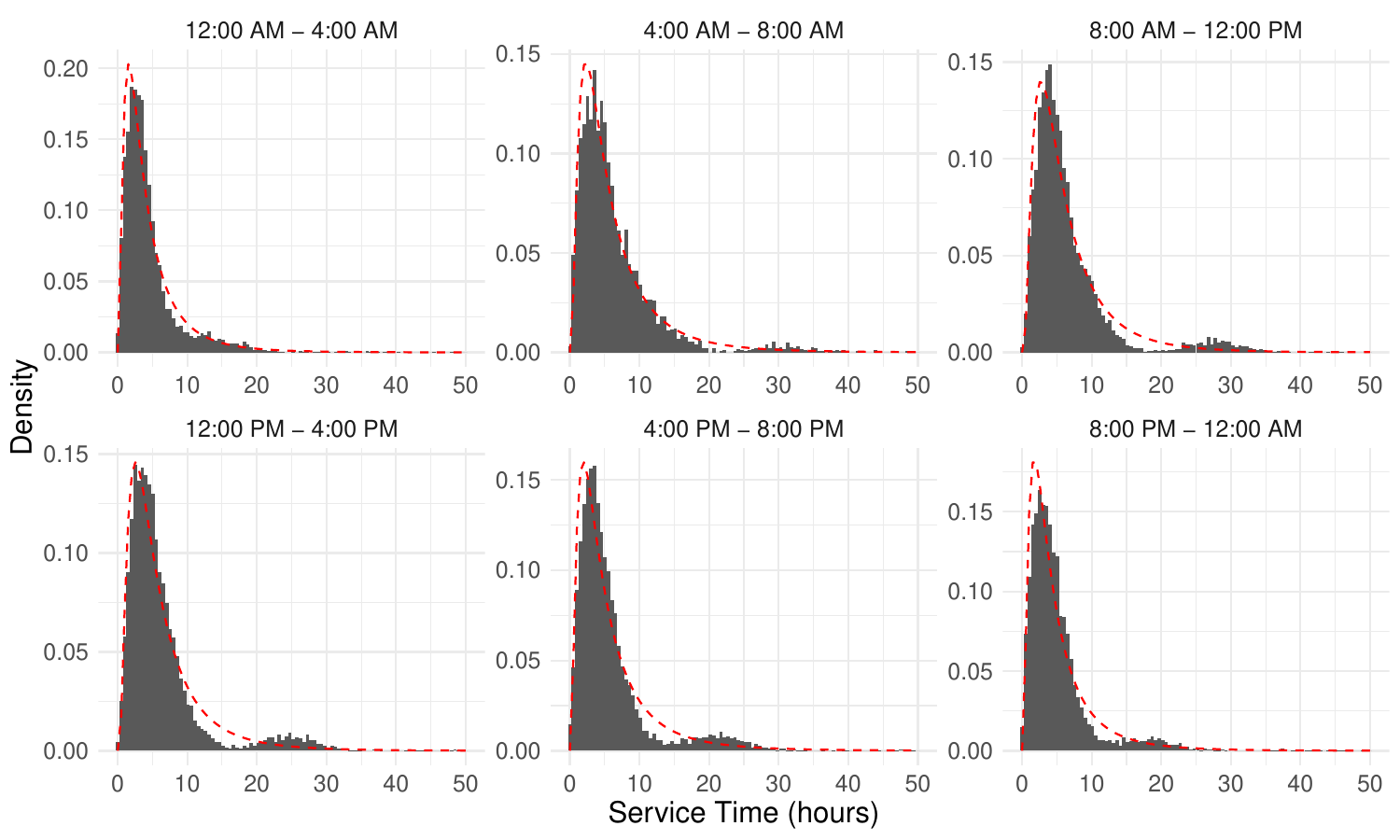}
    \caption{Log-normal density fits (red dashed lines) overlaid on the empirical histograms of service times conditional on time of arrival.}
    \label{fig:servicetime_hist_train}
\end{figure}

Figure~\ref{fig:servicetime_hist_train} displays the histograms of service times of patients upon entering a bed in the ED treatment ward, to until completion of treatment. The histograms are bucketed by time of arrival, and the red dotted lines overlaid onto each plot represent the fitted log-normal distributions.

\section{Experiments} \label{sec:experiments}

\subsection{Patient Satisfaction}
As part of patient satisfaction drive, hospitals are increasingly providing individualized wait time forecasts to patients. However, patients appear to exhibit loss aversion regarding ED wait times, where the discomfort of longer-than-anticipated waits outweighs the gains from shorter-than-expected waits \citep{ansariUnderPromising2022a}. This asymmetry suggests that patients are more satisfied when they are provided with conservative estimates that correspond to a high percentile of their wait time distribution, which \citet{ansariUnderPromising2022a} identified as the 70th percentile. 
Thus we model patient dissatisfaction with the quantile loss
\begin{align*}
    \mathcal{L}_{\alpha}(y, \hat{y}) =
    \begin{cases} 
        \alpha(y - \hat{y}) & \text{if } y - \hat{y} \geq 0 \\
        (1 - \alpha)(\hat{y} - y) & \text{otherwise}
    \end{cases},
\end{align*}
which is minimized by setting the prediction $\hat y$ to the $100\alpha$th percentile of the distribution of $y$. The 70th percentile corresponds to an underage-to-overage cost ratio of $0.7 / (1-0.3) = 2.3$. In Table \ref{tab:qloss_EC} we evaluate how dissatisfied patients are with the percentile predictions from the proposed and benchmark models: Out-of-sample quantile loss values are presented for a range of $\alpha$-values centred around $0.7$. \bgnds consistently reduces patient dissatisfaction by 8-10\% over QRF  across the range of percentiles considered.

We also present the performances of three more approaches. Boosted GLM is the linear boosting approach used in \cite{ansariUnderPromising2022a}. They convert their point estimates to quantiles using a $t$-distribution with homoskedastic standard errors. The classic exponential model specifies the log-rate parameter linearly in the covariates. Finally, we include a naive benchmark called Historical Average, which simply divides a week into 21 bins so that each day of the week consists of three eight-hour bins. Historical average wait time within these bins is used as a point estimate.

\begin{table}
\SingleSpacedXI
\small

\caption{\label{tab:qloss_EC}Out-of-sample quantile loss by target percentiles}
\begin{threeparttable}
\centering
\makebox[\linewidth]
{\begin{tabular}[]{ccccccc}
\toprule
Percentile & Cost Ratio & \bgnds & QRF & Boosted GLM & Classic Exponential & Hist. Avg.\\
\midrule
60 & 1.5 & 0.51 & 0.57 & 0.60 & 0.54 & 0.63 \\
65 & 1.9 & 0.51 & 0.57 & 0.73 & 0.54 & 0.62 \\
70 & 2.3 & 0.50 & 0.55 & 0.95 & 0.53 & 0.60 \\
75 & 3.0 & 0.47 & 0.52 & 1.28 & 0.51 & 0.58 \\
80 & 4.0 & 0.44 & 0.47 & 1.74 & 0.47 & 0.56 \\
\bottomrule
\end{tabular}}
\begin{tablenotes}
\small
\item \textit{Notes: } All figures are rounded to two significant digits.
\end{tablenotes}
\end{threeparttable}
\end{table}

\subsection{Reduction in mortality following cardiac arrest}


For patients suffering from myocardial infarction, \cite{boersmaEarly1996} identified a 2.8 percentage point lower mortality rate when treatment is administered within the \emph{golden hour}, compared with commencing treatment in the second hour. We conduct a stylized analysis where the ED can identify patients who will wait more than 10 minutes and we call these \emph{long waits}. To this end, we focus on patients who showed up with chest pain complaints in our data. We set the 10-minute cutoff assuming the travel time to the ED and conducting tests after the wait time would take up to 50 minutes in total. In this classification task, we employ varying thresholds for flagging a patient as being at-risk of waiting for more than 10 minutes. As an example, if the threshold is 5\%, we would flag patients whose forecasted probability of waiting for more than 10 minutes is amongst the top 5 percent of forecasts. This approach reflects a capacity constraint on the ED, which can only prioritize a finite number of patients.
The top half of Table \ref{tab:mort_EC} presents out-of-sample true positive rates achieved when we threshold the top 5\%, 10\%, ..., 25\% of patient forecasts.

We also present out-of-sample mortality reductions achieved by the proposed and alternative forecasts in the bottom half of Table \ref{tab:mort_EC}. We assume that the ED can reduce mortality rate by 2.8 percentage points if the patient is correctly identified as a \emph{long wait}. Therefore, the overall reduction in mortality rate is the true positive rate multiplied by 0.028. In Table \ref{tab:mort_EC}, we convert the reduction in mortality rate into the number of deaths averted. In the United States, every year, approximately 805,000 people have a heart attack \citep{tsao2022heart}. 


\begin{table}
\SingleSpacedXI
\small

\caption{\label{tab:mort_EC}Identifying long waits among ED visits with chest pain complaints}
\begin{threeparttable}
\centering
\makebox[\linewidth]
{\begin{tabular}[]{cccccc}
\toprule
\% of population & \bgnds & QRF & Boosted GLM & Classic Exponential & Hist. Avg.\\
\midrule
\addlinespace[0.3em]
\multicolumn{6}{c}{\textbf{True positive rate}}\\
\hspace{1em}5 & 0.98 & 0.95 & 0.95 & 0.93 & 0.83 \\
\hspace{1em}10 & 0.97 & 0.94 & 0.94 & 0.92 & 0.83 \\
\hspace{1em}15 & 0.96 & 0.93 & 0.93 & 0.92 & 0.83 \\
\hspace{1em}20 & 0.96 & 0.92 & 0.93 & 0.92 & 0.83 \\
\hspace{1em}25 & 0.95 & 0.92 & 0.92 & 0.92 & 0.83 \\
\addlinespace[0.3em]
\multicolumn{6}{c}{\textbf{Reduction in mortality}}\\
\hspace{1em}5 & 1,100 & 1,100 & 1,100 & 1,000 & 930 \\
\hspace{1em}10 & 2,200 & 2,100 & 2,100 & 2,100 & 1,900\\
\hspace{1em}15 & 3,300 & 3,100 & 3,100 & 3,100 & 2,800\\
\hspace{1em}20 & 4,300 & 4,200 & 4,200 & 4,100 & 3,800\\
\hspace{1em}25 & 5,400 & 5,200 & 5,200 & 5,200 & 4,700\\
\bottomrule
\end{tabular}}
\begin{tablenotes}
\small
\item \textit{Notes: } Identifying patients at-risk of not receiving treatment within the golden hour, so that the ED can prioritize them. All figures are rounded to two significant digits.
\end{tablenotes}
\end{threeparttable}
\end{table}

\end{document}